\documentclass{article}

\makeatletter

\usepackage[nonatbib,preprint]{neurips_2026}

\usepackage[utf8]{inputenc}
\usepackage[T1]{fontenc}
\usepackage{float}
\usepackage{hyperref}
\usepackage{url}
\usepackage{booktabs}
\usepackage{amsfonts}
\usepackage{nicefrac}
\usepackage{microtype}
\usepackage{xcolor}
\usepackage{amsmath}
\usepackage{xspace}
\usepackage{tikz}
\usetikzlibrary{calc}
\usetikzlibrary{shapes.geometric}
\usepackage{pgfplots}
\usepackage{pgfplotstable}
\usepgfplotslibrary{groupplots}
\usepgfplotslibrary{colormaps}
\usepgfplotslibrary{colorbrewer}
\pgfplotsset{compat=1.18}

\newcommand{\constant}{\textsc{Constant}\xspace}
\newcommand{\cooldown}{\textsc{Cooldown}\xspace}
\newcommand{\merge}{\textsc{Merge}\xspace}

\definecolor{constantcolor}{HTML}{d95f02}
\definecolor{cooldowncolor}{HTML}{7570b3}
\definecolor{mergecolor}{HTML}{1b9e77}
\pgfplotsset{
  paper axis/.style={
      /pgfplots/group/xlabels at=edge bottom,
      /pgfplots/group/xticklabels at=edge bottom,
      /pgfplots/group/ylabels at=edge left,
      /pgfplots/group/yticklabels at=edge left,
      axis line style={draw=black!40},
      axis lines*=left,
      every axis plot/.append style={line width=1pt, line join=round},
      every axis title shift=-2pt,
      label style={font=\scriptsize},
      major grid style={draw=black!20},
      minor grid style={draw=black!20},
      tick label style={font=\tiny},
      tick style={draw=black!40},
      title style={font=\scriptsize},
      xmajorgrids,
      ymajorgrids,
    },
  constant plot/.style={
      constantcolor,
      mark=*,
      mark options={
          scale=0.8,
          fill=constantcolor,
          draw=white,
          line width=0.5pt,
        }
    },
  cooldown plot/.style={
      cooldowncolor,
      mark=square*,
      mark options={
          scale=0.72,
          fill=cooldowncolor,
          draw=white,
          line width=0.5pt
        }
    },
  merge plot/.style={
      mergecolor,
      mark=triangle*,
      mark options={
          scale=1.12,
          fill=mergecolor,
          draw=white,
          line width=0.5pt
        }
    },
}

\usepackage[capitalize,noabbrev]{cleveref}
\usepackage[
  backend=biber,
  style=authoryear-comp,
  maxcitenames=1,
  uniquename=false,
  sortcites=ynt,
  uniquelist=false,
  natbib=true,
  hyperref=true,
  maxbibnames=10,
  giveninits=true,
  uniquename=init,
  url=false,
  isbn=false,
  doi=false
]{biblatex}
\DeclareNameAlias{sortname}{family-given}

\title{Good Pretraining, Bad SFT:\\Checkpoint Quality Across the Training Stack}

\author{%
  Sohir Maskey\thanks{Correspondence to \texttt{sohir.maskey@aleph-alpha-research.com}.}
  \quad Philipp Scholl
  \quad Jonas Knupp
  \quad Pit Neitemeier
  \quad Sascha Wirges \\
  Aleph Alpha \\
}

\makeatother

\begin{document}

\maketitle

\begin{figure}[H]
  \centering
  \vspace{-5mm}
  \begin{tikzpicture}
  \newcommand{\evalpanelheight}{0.1\linewidth}
  \newcommand{\evalpanelsep}{0.04\linewidth}
  \begin{groupplot}[
      paper axis,
      group style={
          group size=1 by 2,
          vertical sep=\evalpanelsep,
        },
      scale only axis,
      width=0.42\linewidth,
      height=\evalpanelheight,
      xmin=45,
      xmax=110,
      xtick={60,80,100},
      ymin=-3,
      ymax=40,
      ytick={0,10,20,30,40},
      xlabel={Score retained after perturbation / \%},
      minor x tick num=1,
      ylabel shift=-3pt,
    ]
    \nextgroupplot[
      title={(a) GSM8K},
      legend to name=retentionlegend,
      legend columns=3,
      legend style={draw=none, font=\scriptsize, column sep=0.65em},
    ]
    \addplot[constant plot] coordinates {
        (47.5,0) (52.5,0) (57.5,0) (62.5,0) (67.5,0) (72.5,7) (77.5,13)
        (82.5,18) (87.5,35) (92.5,17) (97.5,8) (102.5,1) (107.5,1)
      };
    \addlegendentry{\constant}
    \addplot[cooldown plot] coordinates {
        (47.5,3) (52.5,4) (57.5,12) (62.5,14) (67.5,16) (72.5,28) (77.5,15)
        (82.5,7) (87.5,1) (92.5,0) (97.5,0) (102.5,0) (107.5,0)
      };
    \addlegendentry{\cooldown}
    \addplot[merge plot] coordinates {
        (47.5,0) (52.5,0) (57.5,0) (62.5,0) (67.5,3) (72.5,2) (77.5,20)
        (82.5,30) (87.5,32) (92.5,13) (97.5,0) (102.5,0) (107.5,0)
      };
    \addlegendentry{\merge}

    \nextgroupplot[title={(b) MBPP}]
    \addplot[constant plot] coordinates {
        (47.5,0) (52.5,0) (57.5,0) (62.5,4) (67.5,4) (72.5,14) (77.5,26)
        (82.5,37) (87.5,12) (92.5,2) (97.5,1) (102.5,0) (107.5,0)
      };
    \addplot[cooldown plot] coordinates {
        (47.5,5) (52.5,6) (57.5,11) (62.5,25) (67.5,27) (72.5,22) (77.5,3)
        (82.5,1) (87.5,0) (92.5,0) (97.5,0) (102.5,0) (107.5,0)
      };
    \addplot[merge plot] coordinates {
        (47.5,1) (52.5,0) (57.5,0) (62.5,4) (67.5,7) (72.5,14) (77.5,32)
        (82.5,29) (87.5,13) (92.5,0) (97.5,0) (102.5,0) (107.5,0)
      };
  \end{groupplot}
  \node[rotate=90, anchor=south, font=\scriptsize]
  at ($($(group c1r1.west)!0.5!(group c1r2.west)$)-(14pt,0)$)
  {Empirical frequency / \%};
  \begin{axis}[
      paper axis,
      at={($(group c1r2.south east)+(0.12\linewidth,0)$)},
      anchor=south west,
      scale only axis,
      width=0.34\linewidth,
      height={2*\evalpanelheight+\evalpanelsep},
      title={(c) Aggregate scores},
      ylabel={Score / \% of best},
      xmin=0.75,
      xmax=4.25,
      ymin=65,
      ymax=105,
      xtick={1,2,3,4},
      xticklabels={Pre,Mid,Long,SFT},
      xlabel={Training stage},
      ytick={70,80,90,100},
      xmajorgrids=false,
      ylabel shift=-3pt,
    ]
    \addplot[constant plot] coordinates {
        (1,90.2) (2,100.0) (3,98.8) (4,99.2)
      };
    \addplot[cooldown plot] coordinates {
        (1,95.7) (2,92.6) (3,98.9) (4,68.0)
      };
    \addplot[merge plot] coordinates {
        (1,100.0) (2,97.1) (3,100.0) (4,100.0)
      };
  \end{axis}
\end{tikzpicture}
\pgfplotslegendfromname{retentionlegend}
   \caption{%
    Perturbation sensitivity and downstream performance of three MoE variants after completing different training stages.
    (a--b) Distribution of scores after 100 Gaussian weight perturbations ($\sigma_\epsilon=0.005$), relative to each checkpoint's unperturbed score; (c) stage aggregates as a percentage of the best checkpoint at that stage.
    \constant and \merge show greater solution density, i.e., perturbing the pretrained checkpoints retains higher scores at performance-preserving thresholds compared to \cooldown.
    This observation is consistent with their stronger post-SFT performance.
  }
  \label{fig:solution_density}
\end{figure}
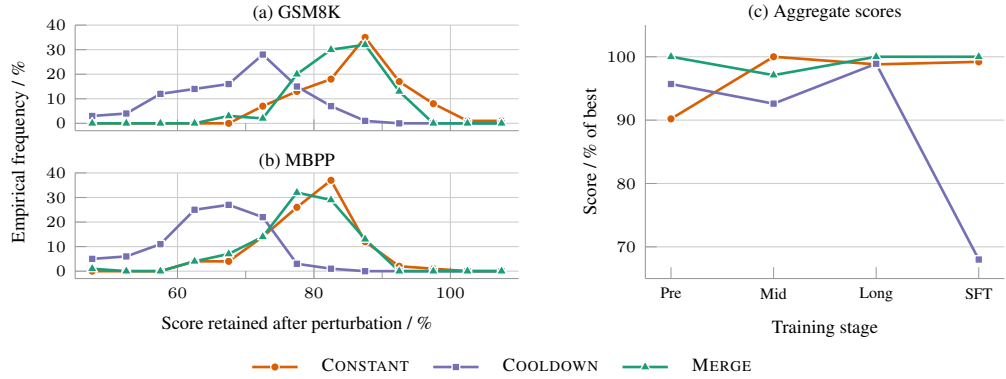

\begin{abstract}
  Language-model checkpoints are commonly selected by pretraining loss or benchmark scores, assuming that the highest-scoring checkpoint will remain the best starting point for subsequent training.
  We show that this assumption can fail in a full 30B mixture-of-experts training pipeline.
  The checkpoints that perform better after the full downstream training stack also have higher \emph{solution density}, i.e., retain downstream performance under local weight perturbations.
\end{abstract}

\section{Introduction}

Language-model checkpoints are usually compared by asking which checkpoint is better \emph{now}: which has lower loss or higher benchmark scores.
This is sufficient when training ends at that checkpoint.
Modern LLMs, however, are trained in a sequence of stages including pretraining, mid-training, long-context adaptation, supervised fine-tuning (SFT), and further post-training \citep{ouyang2022training,chen2023extending,olmo2026olmo3}.
In this setting, the question is different:
\begin{quote}
  \emph{Which checkpoint selection criteria will produce the best final model?}
\end{quote}

This matters because executing all training stages for every candidate is expensive and explodes combinations.
Using intermediate evaluations as selection criteria implicitly assumes that checkpoint rankings are preserved by later training.
Our experiments show that this assumption can fail: subsequent training can reverse the ranking of intermediate checkpoints.

\Cref{tab:intro_flip} compares three source checkpoints from one 30B-parameter mixture-of-experts (MoE) pretraining run.
\constant and \cooldown share the first 6.7T of 7.5T training tokens and differ only in the learning-rate schedule over the final 800B tokens.
\merge is a weighted average of \constant checkpoints \citep{tian2025wsm}.
\cooldown improves every available pretraining signal relative to \constant, yet after identical mid-training, long-context adaptation, and SFT, their ordering reverses.

\begin{table}[t]
  \centering
  \small
  \caption{A better pretraining checkpoint can produce a worse final model.
    \constant and \cooldown share the first 6.7T tokens and the same 7.5T-token budget.
    All three sources then receive the same downstream recipe.
  }
  \label{tab:intro_flip}
  \begin{tabular}{lccc}
    \toprule
    \textbf{Selection signal}        & \textbf{\constant} & \textbf{\cooldown} & \textbf{\merge} \\
    \midrule
    Train loss $\downarrow$          & 1.717              & \textbf{1.627}     & --              \\
    Validation loss $\downarrow$     & 1.734              & \textbf{1.648}     & --              \\
    Pretraining aggregate $\uparrow$ & 0.415              & 0.440              & \textbf{0.460}  \\
    \midrule
    Post-SFT aggregate $\uparrow$    & 0.360              & 0.247              & \textbf{0.363}  \\
    \bottomrule
  \end{tabular}
\end{table}

This motivates our central question: \emph{when do intermediate checkpoint rankings become predictive of the final ranking?}
Across our training trajectories, neither the mid-training nor the long-context aggregate separates \constant from \cooldown, although rankings within a learning-rate sweep stabilize after long-context adaptation (\cref{fig:stage_correlations}).

\citet{gan2026neural} define \emph{solution density} as the fraction of Gaussian perturbations that retain task performance above a threshold.
They show that this fraction increases with model scale, arguing that denser neighborhoods of task-specific experts make useful specialists easier to reach through post-training.
\Cref{fig:solution_density} shows a similar distinction across checkpoints of the same model: \constant and \merge have greater solution density than \cooldown.
\merge is particularly informative: it has the highest pretraining aggregate, remains more robust to perturbations than \cooldown, and finishes effectively tied with \constant after SFT.
This suggests that the unperturbed score alone misses how broadly performance persists under nearby weight changes.
We therefore hypothesize that solution density varies not only with scale but also across training trajectories, and that these differences help explain downstream adaptability.

\paragraph{Contributions.}
We show that conventional pretraining metrics can misrank checkpoints for a fixed downstream training pipeline, and show that intermediate aggregates do not anticipate this reversal.
We further connect this reversal to solution density and show that \cooldown's largest downstream failure reflects unstable response completion rather than absent latent code capability.

\section{Experimental Design}
\label{sec:setup}

We study a 30B-parameter MoE with 3B active parameters.
Training consists of 7.5T-token pretraining, 100B tokens of capability-focused mid-training at 8k context, 100B tokens of long-context adaptation at 64k, and 10B tokens of conversational SFT at 64k.
The optimizer state is reset with repeated LR warmup at each stage.
For details on architecture, training, and datasets see \cref{app:training_details}.

\paragraph{Pretraining sources.}
\constant and \cooldown share their first 6.7T tokens and total token budget.
We decay the learning rate for the \cooldown run to 10\% of the maximum during the final 800B tokens of training.
\merge \citep{tian2025wsm} adds no training: it combines 20 equally spaced \constant checkpoints over a 600B-token trailing window via\footnote{Checkpoint merging is well established and has been ablated in LLM training, for example in Nemotron Super \citep{nvidia2026nemotron3superopen}.
  More directly, \citet{li2025modelmerging} show that averaging constant-LR checkpoints can match annealed pretraining performance, while \citet{tian2025wsm} connect checkpoint weighting to effective update decay.
  See \cref{app:merge_details} for details.
}
\[
  \theta_{\mathrm{merge}}
  =
  \sum_{i=1}^{20}\frac{i}{210}\theta_i,
\]
where the checkpoints are ordered from oldest to newest.

\paragraph{Learning-rate sweep.}
Starting from \cooldown, we train all nine combinations
$
  \eta_{\mathrm{mid}},\eta_{\mathrm{long}}
  \in
  \left\{1,\frac13,\frac19\right\}.
$
where each factor scales a stage's peak learning rate relative to the preceding stage.
For example, $\eta_{\mathrm{long}}=1/3$ sets t4he long-context peak learning rate to one third of the mid-training peak.
Mid-training uses either no decay when $\eta_{\mathrm{long}}=1$ or cosine decay to the long-context peak, yielding a continuous schedule across stages.
\footnote{Following \citet{olmo2026olmo3}, optimizer state is reset and each stage begins with warmup, so continuity holds up to the warmup.}
Long-context training then decays to one third of its peak learning rate.
Stage-wise evaluation aggregates are reported in \cref{tab:fixed_sft3_branches_app}.

The $(1,1)$ mid/long schedule achieved the highest post-SFT aggregate among the nine schedules, each followed by SFT factor $1/3$.
Due to compute constraints, we therefore reuse this schedule for \constant and \merge.
\footnote{This comparison favors \cooldown: its schedule is selected from 9 mid/long configurations, or 27 including the SFT sweep in \cref{app:sft_rates}, whereas \constant and \merge are each run once.
  Tuning them could only improve their scores.
  Thus the reversal's direction in \cref{sec:source_reversal} is unaffected, though its magnitude and the \constant--\merge ordering may change.
  Even the best of 27 \cooldown runs (0.294) trails untuned \constant and \merge (0.360, 0.363).
}

\paragraph{Evaluation.}
At the pretraining, mid-training, and long-context boundaries, we use completion-style benchmarks spanning knowledge, mathematics, and code.
From mid-training onward, we add long-context retrieval.
These three aggregates are cluster-balanced: Pre averages three cluster means, whereas Mid and Long average four cluster means.
Post-SFT, we evaluate on six chat-formatted benchmarks and take their unweighted mean.
The stage-specific suites are given in \cref{app:evaluation_details}.

\section{Checkpoint Rankings Change Across the Training Stack}
\label{sec:results}

\subsection{Pretraining scores do not determine post-SFT scores}
\label{sec:source_reversal}

\Cref{tab:intro_flip} shows that \cooldown improves every available pretraining signal over \constant: train loss (1.717 to 1.627), validation loss (1.734 to 1.648), and pretraining aggregate (0.415 to 0.440).
Under conventional checkpoint selection, \cooldown would therefore dominate \constant.

After the identical mid-training, long-context, and SFT continuation, \constant instead leads \cooldown by 0.113.
\merge has the highest aggregate before and after the downstream pipeline, although its final 0.003 advantage over \constant is small.

\begin{table}[H]
  \centering
  \small
  \setlength{\tabcolsep}{4.2pt}
  \caption{Post-SFT benchmark scores after the fixed downstream recipe; the last column excludes HumanEval+.}
  \label{tab:source_profiles}
  \begin{tabular}{lrrrrrrrr}
    \toprule
    \textbf{Source} & \textbf{AIME} & \textbf{DAPO} & \textbf{Skywork} & \textbf{IFBench} & \textbf{HE+} & \textbf{GPQA} & \textbf{Mean} & \textbf{w/o HE+} \\
    \midrule
    \constant       & 0.221         & 0.430         & 0.390            & 0.238            & 0.646        & 0.237         & 0.360         & 0.303            \\
    \cooldown       & 0.171         & 0.450         & 0.355            & 0.250            & 0.049        & 0.207         & 0.247         & 0.287            \\
    \merge          & 0.217         & 0.495         & 0.345            & 0.236            & 0.665        & 0.222         & 0.363         & 0.303            \\
    \bottomrule
  \end{tabular}
\end{table}

The complete six-task profiles in \cref{tab:source_profiles} show that \cooldown is not uniformly worse.
The large aggregate gap is amplified by a severe HumanEval+ failure, which we examine separately below; without HumanEval+ it shrinks to 0.016 (\cref{tab:source_profiles}).

\subsection{Mid-training rankings remain unstable}

Across the eleven training trajectories, the mid-training aggregate correlates only weakly with the post-SFT aggregate (Pearson $r=0.473$, $p=0.142$; Spearman $\rho=0.482$, $p=0.133$).
\Cref{fig:solution_density}c and \cref{fig:stage_correlations} show that rankings can still change substantially after this stage.

\subsection{Long-context rankings stabilize within a sweep, not across sources}

After long-context adaptation, the association changes substantially.
Pearson correlation with the post-SFT aggregate rises to $r=0.884$ ($p<0.001$) and Spearman correlation to $\rho=0.964$ ($p<0.001$).

\begin{figure}[t]
  \centering
  \begin{tikzpicture}
  \begin{groupplot}[
      paper axis,
      group style={group size=2 by 1, horizontal sep=0.055\linewidth},
      width=0.49\linewidth,
      height=0.26\linewidth,
      ymin=0.05,
      ymax=0.39,
      ytick={0.1,0.2,0.3},
      every axis plot/.append style={every mark/.append style={scale=1.4}},
    ]
    \nextgroupplot[
      title={(a) After mid-training},
      xmin=0.505,
      xmax=0.570,
      xtick={0.51,0.53,0.55,0.57},
      xlabel={Mid-training aggregate},
      ylabel={Post-SFT aggregate},
      legend to name=stagecorrelationlegend,
      legend columns=3,
      legend style={draw=none, font=\scriptsize, column sep=0.7em},
    ]
    \addplot[black!55, dashed, forget plot] coordinates {
        (0.505,0.1292) (0.570,0.2832)
      };
    \addplot[constant plot, only marks, every mark/.append style={scale=1.4}] coordinates {
        (0.555811,0.360441)
      };
    \addlegendentry{\constant}
    \addplot[cooldown plot, only marks, every mark/.append style={scale=1.4}] coordinates {
        (0.514992,0.246947)
        (0.563698,0.198606)
        (0.544467,0.186825)
        (0.556671,0.228053)
        (0.543338,0.190754)
        (0.536744,0.118977)
        (0.515561,0.160391)
        (0.520513,0.104497)
        (0.511721,0.086749)
      };
    \addlegendentry{\cooldown}
    \addplot[merge plot, only marks, every mark/.append style={scale=1.4}] coordinates {
        (0.539504,0.363320)
      };
    \addlegendentry{\merge}
    \node[font=\tiny, anchor=north west, fill=white, fill opacity=0.82,
      text opacity=1, inner sep=1.5pt]
    at (rel axis cs:0.03,0.97) {$r=0.473,\ \rho=0.482$};

    \nextgroupplot[
      title={(b) After long-context adaptation},
      xmin=0.575,
      xmax=0.650,
      xtick={0.58,0.60,0.62,0.64},
      xlabel={Long-context aggregate},
      yticklabels={},
    ]
    \addplot[black!55, dashed, forget plot] coordinates {
        (0.575,0.0330) (0.650,0.3363)
      };
    \addplot[constant plot, only marks, every mark/.append style={scale=1.4}] coordinates {
        (0.636204,0.360441)
      };
    \addplot[cooldown plot, only marks, every mark/.append style={scale=1.4}] coordinates {
        (0.637336,0.246947)
        (0.626483,0.198606)
        (0.621465,0.186825)
        (0.625964,0.228053)
        (0.617950,0.190754)
        (0.610240,0.118977)
        (0.601995,0.160391)
        (0.588271,0.104497)
        (0.580863,0.086749)
      };
    \addplot[merge plot, only marks, every mark/.append style={scale=1.4}] coordinates {
        (0.643691,0.363320)
      };
    \node[font=\tiny, anchor=north west, fill=white, fill opacity=0.82,
      text opacity=1, inner sep=1.5pt]
    at (rel axis cs:0.03,0.97) {$r=0.884,\ \rho=0.964$};
  \end{groupplot}
\end{tikzpicture}
\pgfplotslegendfromname{stagecorrelationlegend}
   \caption{Intermediate versus post-SFT aggregate scores.
    Dashed lines are ordinary least-squares fits.
  }
  \label{fig:stage_correlations}
\end{figure}
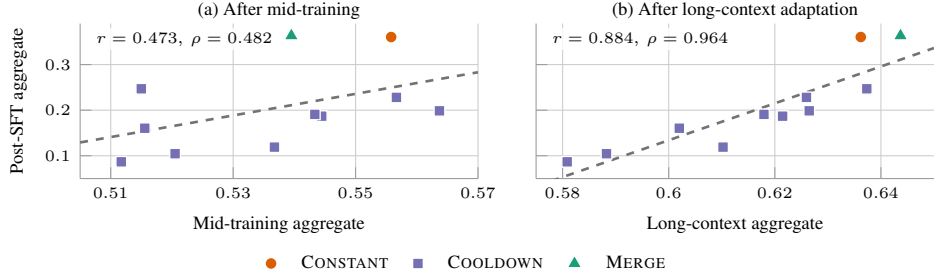

\Cref{fig:stage_correlations} compares the post-SFT aggregate against two intermediate aggregates.
The diffuse mid-training relationship tightens into an almost monotone ordering after long-context adaptation.

This result is not driven by adding \constant and \merge: restricting the calculation to the nine \cooldown trajectories gives $r=0.448$ ($p=0.226$), $\rho=0.450$ ($p=0.224$) after mid-training and $r=0.935$, $\rho=0.950$ (both $p<0.001$) after long-context adaptation.
Notably, long-context scores still fail to predict the ordering of \constant, \cooldown, and \merge at matched schedules: \cooldown $(1,1)$ scores 0.637 against 0.636 for \constant, yet trails by 0.113 after SFT.

We do not interpret this as evidence that long-context adaptation causally creates predictiveness: later checkpoints are also closer to the final model.

\section{What Checkpoint Scores Miss}
\label{sec:diagnostics}

The ranking reversal suggests that checkpoint scores miss properties that matter for subsequent behavior.
We examine this in two ways.

\subsection{Solution density separates the more trainable sources}

The preceding results define \emph{trainability} operationally: how well a source checkpoint performs after the same remaining training pipeline.
But what local parameter-space geometry provides a signal of this pipeline-conditioned quality that the unperturbed checkpoint score misses?

For checkpoint $\theta$, benchmark $b$, perturbation $\epsilon$, and evaluation score $s_b(\theta)$, we define the relative-threshold solution-density profile following \citet{gan2026neural} as
\begin{equation}
  \delta_{\theta,b}(\tau)
  =
  \Pr_{\epsilon}
  \left[
    s_b(\theta+\epsilon)\geq\tau s_b(\theta)
    \right].
\end{equation}

Each source checkpoint is evaluated under 100 Gaussian perturbations on GSM8K and MBPP at standard deviations $\sigma_\epsilon\in\{0.005,0.001\}$.
\cref{fig:solution_density} shows that \cooldown is shifted toward lower retained performance on both tasks.
Thresholded profiles and exact counts appear in \cref{fig:solution_density_primary,tab:retention_counts}.
On GSM8K, solution density at $\tau=0.90$ is 27\% for \constant and 13\% for \merge, but 0\% for \cooldown.
The separation persists at $\sigma_\epsilon=0.001$ (\cref{fig:solution_density_small,tab:retention_counts_small}).

\begin{figure}[t]
  \centering
\pgfplotsset{
  colormap={accuracychange}{
      rgb255(0cm)=(49,54,149)
      rgb255(0.3cm)=(69,117,180)
      rgb255(0.47cm)=(146,197,222)
      rgb255(0.5cm)=(247,247,247)
      rgb255(0.54cm)=(254,224,139)
      rgb255(0.64cm)=(239,101,72)
      rgb255(1cm)=(165,0,38)
    },
}
\begin{tikzpicture}
  \newcommand{\umappanelwidth}{0.2\linewidth}
  \newcommand{\umappanelheight}{\umappanelwidth}  %
  \newcommand{\umappanelsep}{0.025\linewidth}
  \newcommand{\umapstar}[2]{
    \node[star, star points=5, star point ratio=2.5, fill=black,
      draw=white, line width=0.75pt, inner sep=1.2pt]
    at (axis cs:#1,#2,2) {};
  }
  \newcommand{\umapbase}{
    \node[circle, fill=black, minimum size=2.2pt, inner sep=0pt]
    at (axis cs:0.5,0.5,2) {};
  }
  \begin{groupplot}[
      paper axis,
      group style={
          group size=3 by 2,
          horizontal sep=\umappanelsep,
          vertical sep=\umappanelsep,
        },
      scale only axis,
      width=\umappanelwidth,
      height=\umappanelheight,
      axis lines=box,
      enlargelimits=false,
      xtick=\empty,
      ytick=\empty,
      colormap name=accuracychange,
      point meta min=0,
      point meta max=1,
      view={0}{90},
      ylabel shift=4pt,
    ]
    \nextgroupplot[ylabel={GSM8K}]
    \addplot3[surf, shader=interp, mesh/rows=81, mesh/cols=81] table[col sep=comma, x=x, y=y, z=z] {figures/umap_grid/gsm8k_constant.csv};
    \umapbase
    \umapstar{0.7000}{0.4500}
    \nextgroupplot
    \addplot3[surf, shader=interp, mesh/rows=81, mesh/cols=81] table[col sep=comma, x=x, y=y, z=z] {figures/umap_grid/gsm8k_cooldown.csv};
    \umapbase
    \umapstar{0.3750}{0.4875}
    \nextgroupplot[
      colorbar,
      colorbar style={
          height=\umappanelheight,
          tick label style={font=\tiny},
          label style={font=\scriptsize},
          ylabel={Accuracy change / \%},
          ytick={0.0000,0.5000,1.0000},
          yticklabels={-58,0,10},
        },
    ]
    \addplot3[surf, shader=interp, mesh/rows=81, mesh/cols=81] table[col sep=comma, x=x, y=y, z=z] {figures/umap_grid/gsm8k_merge.csv};
    \umapbase
    \umapstar{0.6250}{0.5375}
    \nextgroupplot[xlabel={\constant}, ylabel={MBPP}]
    \addplot3[surf, shader=interp, mesh/rows=81, mesh/cols=81] table[col sep=comma, x=x, y=y, z=z] {figures/umap_grid/mbpp_constant.csv};
    \umapbase
    \umapstar{0.5125}{0.1125}
    \nextgroupplot[xlabel={\cooldown}]
    \addplot3[surf, shader=interp, mesh/rows=81, mesh/cols=81] table[col sep=comma, x=x, y=y, z=z] {figures/umap_grid/mbpp_cooldown.csv};
    \umapbase
    \umapstar{0.4875}{0.5000}
    \nextgroupplot[
      xlabel={\merge},
      colorbar,
      colorbar style={
          height=\umappanelheight,
          tick label style={font=\tiny},
          label style={font=\scriptsize},
          ylabel={Accuracy change / \%},
          ytick={0.0000,0.5000,1.0000},
          yticklabels={-74,0,8},
        },
    ]
    \addplot3[surf, shader=interp, mesh/rows=81, mesh/cols=81] table[col sep=comma, x=x, y=y, z=z] {figures/umap_grid/mbpp_merge.csv};
    \umapbase
    \umapstar{0.5500}{0.4250}
  \end{groupplot}
\end{tikzpicture}
   \caption{
    Filled contours show interpolated accuracy changes relative to unperturbed checkpoints on GSM8K (top) and MBPP (bottom).
    Stars mark each panel's optimum.
    \constant has the greatest solution density, whereas \merge starts stronger and degrades less under perturbation than \cooldown, especially on MBPP (\Cref{tab:retention_counts_small}).
    Projections are descriptive only.
  }
  \label{fig:perturbation_umap}
\end{figure}
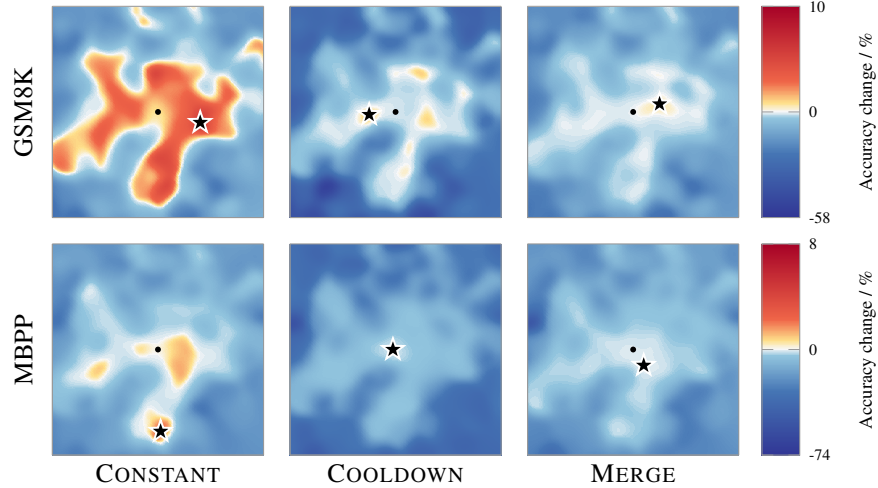

\Cref{fig:perturbation_umap} provides a complementary two-dimensional view of the perturbation outcomes.

The perturbation ordering matches the controlled trainability result in \cref{sec:source_reversal,tab:source_profiles}: \constant and \merge finish nearly tied at 0.360 and 0.363, while \cooldown reaches 0.247 despite its stronger pretraining signals (\cref{tab:intro_flip}).
We hypothesize that greater local robustness helps these checkpoints tolerate the parameter displacement induced by subsequent optimization.

\subsection{Retuning SFT does not repair stopping}

\cooldown's largest deficit is HumanEval+ (4.9\% versus 64.6\% and 66.5\% for \constant and \merge).
Raw outputs reveal a broader failure: nearly all AIME and GPQA responses reach the 32k-token cap, with highly repetitive tails.
Because this pathology appears after SFT, we hypothesized that the SFT stage was responsible and swept its learning rate.

The sweep did not repair stopping.
Changing the SFT factor from $1/3$ to $1$ leaves 100\% of AIME and 99\% of GPQA responses at the cap, and although it raises the post-SFT aggregate from 0.247 to 0.294, no tested rate jointly resolves the failure: lower rates improve HumanEval+ on average but reduce reasoning and the overall aggregate.

Sweeping all three SFT factors for each of the nine \cooldown mid/long combinations leaves all 27 checkpoints below the fixed-recipe \constant and \merge checkpoints.
Evaluating only the first generated code block from one \cooldown checkpoint raises HumanEval+ from 7.3\% to 61.0\%, showing latent code capability but still trailing \constant and \merge (64.6\% and 66.5\%), though thousands of repeated later blocks remain a genuine failure.
\Cref{app:sft_rates} summarizes the nested sweep with additional trajectory results, traces, and extraction details.

\section{Related Work}
\label{sec:related_work}

Prior work shows that pretraining loss need not predict downstream adaptation: models with matched loss can transfer differently \citep{liu2023same}, and learning-rate decay can improve pretraining metrics while hurting continued training and SFT \citep{yano2026pretraining}.
Related work links flatter LLM solutions to better trainability \citep{li2024forgetting,watts2026sharpness}.
We extend this line by tracking checkpoint rankings across multiple training stages and asking when they predict the final ranking.
We also build on solution density, which measures whether nearby perturbations retain task performance \citep{gan2026neural}, and on checkpoint merging, where checkpoints induce an effective decay over updates \citep{li2025modelmerging,tian2025wsm}.
Extended related work appears in \cref{app:related_work}.

\section{Conclusion}
\label{sec:discussion}

Checkpoint quality depends on what comes next.
\cooldown has better pretraining metrics than \constant but performs worse after the downstream pipeline.
Rankings within a learning-rate sweep stabilize after long-context adaptation, yet neither intermediate aggregate anticipates the reversal between \constant and \cooldown.
Our audits further show that eval scores can miss properties relevant to continued training.
Checkpoint selection should therefore target performance after the remaining pipeline, not the current score alone.

\paragraph{Limitations.}
Our results come from one 30B MoE family with one seed per setup, and we did not test every combination of settings.
Solution density is measured on only two tasks and does not establish causation, and \cooldown's stopping failure enlarges the observed gap.

\begingroup
\emergencystretch=2em
\sloppy
\printbibliography

@inproceedings{vaswani2017attention,
  title         = {Attention Is All You Need},
  author        = {Vaswani, Ashish and Shazeer, Noam and Parmar, Niki and Uszkoreit, Jakob and Jones, Llion and Gomez, Aidan N. and Kaiser, Lukasz and Polosukhin, Illia},
  booktitle     = {Advances in Neural Information Processing Systems},
  volume        = {30},
  year          = {2017}
}

@inproceedings{ainslie2023gqa,
  title         = {{GQA}: Training Generalized Multi-Query Transformer Models from Multi-Head Checkpoints},
  author        = {Ainslie, Joshua and Lee-Thorp, James and de Jong, Michiel and Zemlyanskiy, Yury and Lebr{\'o}n, Federico and Sanghai, Sumit},
  booktitle     = {Proceedings of the 2023 Conference on Empirical Methods in Natural Language Processing},
  year          = {2023}
}

@misc{su2021roformer,
  title         = {{RoFormer}: Enhanced Transformer with Rotary Position Embedding},
  author        = {Su, Jianlin and Lu, Yu and Pan, Shengfeng and Murtadha, Ahmed and Wen, Bo and Liu, Yunfeng},
  year          = {2021},
  eprint        = {2104.09864},
  archiveprefix = {arXiv}
}

@inproceedings{zhang2019rmsnorm,
  title         = {Root Mean Square Layer Normalization},
  author        = {Zhang, Biao and Sennrich, Rico},
  booktitle     = {Advances in Neural Information Processing Systems},
  volume        = {32},
  year          = {2019}
}

@misc{shazeer2020glu,
  title         = {{GLU} Variants Improve Transformer},
  author        = {Shazeer, Noam},
  year          = {2020},
  eprint        = {2002.05202},
  archiveprefix = {arXiv}
}

@inproceedings{loshchilov2019decoupled,
  title         = {Decoupled Weight Decay Regularization},
  author        = {Loshchilov, Ilya and Hutter, Frank},
  booktitle     = {International Conference on Learning Representations},
  year          = {2019}
}

@misc{jordan2024muon,
  title         = {Muon: An Optimizer for Hidden Layers in Neural Networks},
  author        = {Jordan, Keller and Jin, Yuchen and Boza, Vlado and You, Jiacheng and Cesista, Franz and Newhouse, Laker and Bernstein, Jeremy},
  year          = {2024},
  url           = {https://kellerjordan.github.io/posts/muon/}
}

@misc{liang2024torchtitan,
  title         = {{TorchTitan}: One-Stop {PyTorch} Native Solution for Production-Ready {LLM} Pretraining},
  author        = {Liang, Wanchao and Liu, Tianyu and Wright, Less and Constable, Will and Gu, Andrew and Huang, Chien-Chin and Zhang, Iris and Feng, Wei and Huang, Howard and Wang, Junjie and others},
  year          = {2024},
  eprint        = {2410.06511},
  archiveprefix = {arXiv}
}

@misc{gan2026neural,
  title         = {Neural Thickets: Diverse Task Experts Are Dense Around Pretrained Weights},
  author        = {Gan, Yulu and Isola, Phillip},
  year          = {2026},
  eprint        = {2603.12228},
  archiveprefix = {arXiv},
  primaryclass  = {cs.LG},
  url           = {https://arxiv.org/abs/2603.12228}
}

@article{chen2023extending,
  title         = {Extending Context Window of Large Language Models via Positional Interpolation},
  author        = {Chen, Shouyuan and Wong, Sherman and Chen, Liangjian and Tian, Yuandong},
  journal       = {arXiv preprint arXiv:2306.15595},
  year          = {2023}
}

@article{ouyang2022training,
  title         = {Training Language Models to Follow Instructions with Human Feedback},
  author        = {Ouyang, Long and Wu, Jeffrey and Jiang, Xu and Almeida, Diogo and Wainwright, Carroll L. and Mishkin, Pamela and Zhang, Chong and Agarwal, Sandhini and Slama, Katarina and Ray, Alex and others},
  journal       = {Advances in Neural Information Processing Systems},
  volume        = {35},
  pages         = {27730--27744},
  year          = {2022}
}

@misc{olmo2026olmo3,
  title         = {Olmo 3},
  author        = {Team Olmo and : and Allyson Ettinger and Amanda Bertsch and Bailey Kuehl and David Graham and David Heineman and Dirk Groeneveld and Faeze Brahman and Finbarr Timbers and Hamish Ivison and Jacob Morrison and Jake Poznanski and Kyle Lo and Luca Soldaini and Matt Jordan and Mayee Chen and Michael Noukhovitch and Nathan Lambert and Pete Walsh and Pradeep Dasigi and Robert Berry and Saumya Malik and Saurabh Shah and Scott Geng and Shane Arora and Shashank Gupta and Taira Anderson and Teng Xiao and Tyler Murray and Tyler Romero and Victoria Graf and Akari Asai and Akshita Bhagia and Alexander Wettig and Alisa Liu and Aman Rangapur and Chloe Anastasiades and Costa Huang and Dustin Schwenk and Harsh Trivedi and Ian Magnusson and Jaron Lochner and Jiacheng Liu and Lester James V. Miranda and Maarten Sap and Malia Morgan and Michael Schmitz and Michal Guerquin and Michael Wilson and Regan Huff and Ronan Le Bras and Rui Xin and Rulin Shao and Sam Skjonsberg and Shannon Zejiang Shen and Shuyue Stella Li and Tucker Wilde and Valentina Pyatkin and Will Merrill and Yapei Chang and Yuling Gu and Zhiyuan Zeng and Ashish Sabharwal and Luke Zettlemoyer and Pang Wei Koh and Ali Farhadi and Noah A. Smith and Hannaneh Hajishirzi},
  year          = {2026},
  eprint        = {2512.13961},
  archiveprefix = {arXiv},
  primaryclass  = {cs.CL},
  url           = {https://arxiv.org/abs/2512.13961}
}

@inproceedings{clark2018arc,
  title         = {Think You Have Solved Question Answering? Try {ARC}, the {AI2} Reasoning Challenge},
  author        = {Clark, Peter and Cowhey, Isaac and Etzioni, Oren and Khot, Tushar and Sabharwal, Ashish and Schoenick, Carissa and Tafjord, Oyvind},
  booktitle     = {arXiv preprint arXiv:1803.05457},
  year          = {2018}
}

@inproceedings{zellers2019hellaswag,
  title         = {{HellaSwag}: Can a Machine Really Finish Your Sentence?},
  author        = {Zellers, Rowan and Holtzman, Ari and Bisk, Yonatan and Farhadi, Ali and Choi, Yejin},
  booktitle     = {Proceedings of the 57th Annual Meeting of the Association for Computational Linguistics},
  year          = {2019}
}

@inproceedings{hendrycks2020mmlu,
  title         = {Measuring Massive Multitask Language Understanding},
  author        = {Hendrycks, Dan and Burns, Collin and Basart, Steven and Zou, Andy and Mazeika, Mantas and Song, Dawn and Steinhardt, Jacob},
  booktitle     = {International Conference on Learning Representations (ICLR)},
  year          = {2021}
}

@misc{wang2024mmlupro,
  title         = {{MMLU-Pro}: A More Robust and Challenging Multi-Task Language Understanding Benchmark},
  author        = {Wang, Yubo and Ma, Xueguang and Zhang, Ge and Ni, Yuansheng and Chandra, Abhranil and Guo, Shiguang and Ren, Weiming and Arulraj, Aaran and He, Xuan and Jiang, Ziyan and others},
  year          = {2024},
  eprint        = {2406.01574},
  archiveprefix = {arXiv}
}

@inproceedings{bisk2019piqa,
  title         = {{PIQA}: Reasoning about Physical Commonsense in Natural Language},
  author        = {Bisk, Yonatan and Zellers, Rowan and Le bras, Ronan and Gao, Jianfeng and Choi, Yejin},
  booktitle     = {Proceedings of the AAAI Conference on Artificial Intelligence},
  year          = {2020}
}

@inproceedings{joshi2017triviaqa,
  title         = {{TriviaQA}: A Large Scale Distantly Supervised Challenge Dataset for Reading Comprehension},
  author        = {Joshi, Mandar and Choi, Eunsol and Weld, Daniel and Zettlemoyer, Luke},
  booktitle     = {Proceedings of the 55th Annual Meeting of the Association for Computational Linguistics},
  year          = {2017}
}

@misc{cobbe2021gsm8k,
  title         = {Training Verifiers to Solve Math Word Problems},
  author        = {Cobbe, Karl and Kosaraju, Vineet and Bavarian, Mohammad and Chen, Mark and Jun, Heewoo and Kaiser, Lukasz and Plappert, Matthias and Tworek, Jerry and Hilton, Jacob and Nakano, Rei and Hesse, Christopher and Schulman, John},
  year          = {2021},
  eprint        = {2110.14168},
  archiveprefix = {arXiv}
}

@misc{hendrycks2021math,
  title         = {Measuring Mathematical Problem Solving With the {MATH} Dataset},
  author        = {Hendrycks, Dan and Burns, Collin and Kadavath, Saurav and Arora, Akul and Basart, Steven and Tang, Eric and Song, Dawn and Steinhardt, Jacob},
  year          = {2021},
  eprint        = {2103.03874},
  archiveprefix = {arXiv}
}

@misc{chen2021humaneval,
  title         = {Evaluating Large Language Models Trained on Code},
  author        = {Chen, Mark and Tworek, Jerry and Jun, Heewoo and Yuan, Qiming and Pinto, Henrique Ponde de Oliveira and Kaplan, Jared and Edwards, Harri and Burda, Yuri and Joseph, Nicholas and Brockman, Greg and others},
  year          = {2021},
  eprint        = {2107.03374},
  archiveprefix = {arXiv}
}

@misc{austin2021program,
  title         = {Program Synthesis with Large Language Models},
  author        = {Austin, Jacob and Odena, Augustus and Nye, Maxwell and Bosma, Maarten and Michalewski, Henryk and Dohan, David and Jiang, Ellen and Cai, Carrie and Terry, Michael and Le, Quoc and Sutton, Charles},
  year          = {2021},
  eprint        = {2108.07732},
  archiveprefix = {arXiv}
}

@misc{hsieh2024ruler,
  title         = {{RULER}: What's the Real Context Size of Your Long-Context Language Models?},
  author        = {Hsieh, Cheng-Ping and Sun, Simeng and Kriman, Samuel and Acharya, Shantanu and Rekesh, Dima and Jia, Fei and Ginsburg, Boris},
  year          = {2024},
  eprint        = {2404.06654},
  archiveprefix = {arXiv}
}

@misc{yen2024helmet,
  title         = {{HELMET}: How to Evaluate Long-Context Language Models Effectively and Thoroughly},
  author        = {Yen, Howard and Gao, Tianyu and Hou, Minmin and Ding, Ke and Fleischer, Daniel and Izsak, Peter and Wasserblat, Moshe and Chen, Danqi},
  year          = {2024},
  eprint        = {2410.02694},
  archiveprefix = {arXiv}
}

@misc{yu2025dapo,
  title         = {{DAPO}: An Open-Source {LLM} Reinforcement Learning System at Scale},
  author        = {Yu, Qiying and Zhang, Zheng and Zhu, Ruofei and Yuan, Yufeng and Zuo, Xiaochen and Yue, Yu and Dai, Weinan and Fan, Tiantian and Liu, Gaohong and Liu, Lingjun and others},
  year          = {2025},
  eprint        = {2503.14476},
  archiveprefix = {arXiv}
}

@misc{he2025skywork,
  title         = {Skywork Open Reasoner 1 Technical Report},
  author        = {He, Jujie and Liu, Jiacai and Liu, Chris Yuhao and Yan, Rui and Wang, Chaojie and Cheng, Peng and Zhang, Xiaoyu and Zhang, Fuxiang and Xu, Jiacheng and Shen, Wei and others},
  year          = {2025},
  eprint        = {2505.22312},
  archiveprefix = {arXiv}
}

@inproceedings{pyatkin2025ifbench,
  title         = {Generalizing Verifiable Instruction Following},
  author        = {Pyatkin, Valentina and Malik, Saumya and Graf, Victoria and Ivison, Hamish and Huang, Shengyi and Dasigi, Pradeep and Lambert, Nathan and Hajishirzi, Hannaneh},
  booktitle     = {Advances in Neural Information Processing Systems},
  year          = {2025}
}

@misc{liu2023evalplus,
  title         = {Is Your Code Generated by {ChatGPT} Really Correct? Rigorous Evaluation of Large Language Models for Code Generation},
  author        = {Liu, Jiawei and Xia, Chunqiu Steven and Wang, Yuyao and Zhang, Lingming},
  year          = {2023},
  eprint        = {2305.01210},
  archiveprefix = {arXiv}
}

@misc{rein2023gpqa,
  title         = {{GPQA}: A Graduate-Level Google-Proof Q\&A Benchmark},
  author        = {Rein, David and Hou, Betty Li and Stickland, Asa Cooper and Petty, Jackson and Pang, Richard Yuanzhe and Dirani, Julien and Michael, Julian and Bowman, Samuel R.},
  year          = {2023},
  eprint        = {2311.12022},
  archiveprefix = {arXiv}
}

@misc{nvidia2026nemotron3superopen,
  title         = {Nemotron 3 Super: Open, Efficient Mixture-of-Experts Hybrid Mamba-Transformer Model for Agentic Reasoning},
  author        = {NVIDIA and : and Aakshita Chandiramani and Aaron Blakeman and Abdullahi Olaoye and Abhibha Gupta and Abhilash Somasamudramath and Abhinav Khattar and Adeola Adesoba and Adi Renduchintala and Adil Asif and Aditya Agrawal and Aditya Vavre and Ahmad Kiswani and Aishwarya Padmakumar and Ajay Hotchandani and Akanksha Shukla and Akhiad Bercovich and Aleksander Ficek and Aleksandr Shaposhnikov and Alex Gronskiy and Alex Kondratenko and Alex Neefus and Alex Steiner and Alex Yang and Alexander Bukharin and Alexander Young and Ali Hatamizadeh and Ali Taghibakhshi and Alina Galiautdinova and Alisa Liu and Alok Kumar and Ameya Sunil Mahabaleshwarkar and Amir Klein and Amit Zuker and Amnon Geifman and Anahita Bhiwandiwalla and Ananth Subramaniam and Andrew Tao and Anjaney Shrivastava and Anjulie Agrusa and Ankur Srivastava and Ankur Verma and Ann Guan and Anna Shors and Annamalai Chockalingam and Anubhav Mandarwal and Aparnaa Ramani and Arham Mehta and Arti Jain and Arun Venkatesan and Asha Anoosheh and Ashwath Aithal and Ashwin Poojary and Asif Ahamed and Asit Mishra and Asli Sabanci Demiroz and Asma Kuriparambil Thekkumpate and Atefeh Sohrabizadeh and Avinash Kaur and Ayush Dattagupta and Barath Subramaniam Anandan and Bardiya Sadeghi and Barnaby Simkin and Ben Lanir and Benedikt Schifferer and Benjamin Chislett and Besmira Nushi and Bilal Kartal and Bill Thiede and Bita Darvish Rouhani and Bobby Chen and Boris Ginsburg and Brandon Norick and Branislav Kisacanin and Brian Yu and Bryan Catanzaro and Buvaneswari Mani and Carlo del Mundo and Chankyu Lee and Chanran Kim and Chantal Hwang and Chao Ni and Charles Wang and Charlie Truong and Cheng-Ping Hsieh and Chenhan Yu and Chenjie Luo and Cherie Wang and Chetan Mungekar and Chintan Patel and Chris Alexiuk and Chris Holguin and Chris Wing and Christian Munley and Christopher Parisien and Chuck Desai and Chunyang Sheng and Collin Neale and Cyril Meurillon and Dakshi Kumar and Dan Gil and Dan Su and Dane Corneil and Daniel Afrimi and Daniel Burkhardt Eliuth Triana and Daniel Egert and Daniel Fatade and Daniel Lo and Daniel Rohrer and Daniel Serebrenik and Daniil Sorokin and Daria Gitman and Daria Levy and Darko Stosic and David Edelsohn and David Messina and David Mosallanezhad and David Tamok and Deena Donia and Deepak Narayanan and Devin O'Kelly and Dheeraj Peri and Dhruv Nathawani and Di Wu and Dima Rekesh and Dina Yared and Divyanshu Kakwani and Dmitry Konyagin Brandon Tuttle and Dong Ahn and Dongfu Jiang and Dorrin Poorkay and Douglas O'Flaherty and Duncan Riach and Dusan Stosic and Dustin Van Stee and Edgar Minasyan and Edward Lin and Eileen Peters Long and Elad Segal and Elena Lantz and Elena Lewis and Ellie Evans and Elliott Ning and Eric Chung and Eric Harper and Eric Pham-Hung and Eric W. Tramel and Erick Galinkin and Erik Pounds and Esti Etrog and Evan Briones and Evan Wu and Evelina Bakhturina and Evgeny Tsykunov and Ewa Dobrowolska and Farshad Saberi Movahed and Farzan Memarian and Fay Wang and Fei Jia and Felipe Soares and Felipe Vieira Frujeri and Feng Chen and Fengguang Lin and Ferenc Galko and Fortuna Zhang and Frankie Siino and Frida Hou and Gantavya Bhatt and Gargi Prasad and Geethapriya Venkataramani and Geetika Gupta and George Armstrong and Gerald Shen and Giulio Borghesi and Gordana Neskovic and Gorkem Batmaz and Grace Lam and Grace Wu and Greg Pauloski and Greyson Davis and Grigor Nalbandyan and Guoming Zhang and Guy Farber and Guyue Huang and Haifeng Qian and Haran Kumar Shiv Kumar and Harry Kim and Harsh Sharma and Hayate Iso and Hayley Ross and Herbert Hum and Herman Sahota and Hexin Wang and Himanshu Soni and Hiren Upadhyay and Huy Nguyen and Iain Cunningham and Ido Galil and Ido Shahaf and Igino Padovani and Igor Gitman and Igor Shovkun and Ikroop Dhillon and Ilya Loshchilov and Ingrid Kelly and Itamar Schen and Itay Levy and Ivan Moshkov and Izik Golan and Izzy Putterman and Jain Tu and Jan Baczek and Jan Kautz and Jane Polak Scowcroft and Janica Rosenberg and Jared Casper and Jarrod Pflum and Jason Grant and Jason Sewall and Jatin Mitra and Jeffrey Glick and Jenny Chen and Jesse Oliver and Jiacheng Xu and Jiafan Zhu and Jialin Song and Jian Zhang and Jiaqi Zeng and Jie Lou and Jill Milton and Jim Chow and Jimmy Zhang and Jinhang Choi and Jining Huang and Jocelyn Huang and Joel Caruso and Joey Conway and Joey Guman and Johan Jatko and John Kamalu and Johnny Greco and Jonathan Cohen and Jonathan Raiman and Joseph Jennings and Joyjit Daw and Juan Yu and Julio Tapia and Junkeun Yi and Jupinder Parmar and Jyothi Achar and Kari Briski and Kartik Mattoo and Katherine Cheung and Katherine Luna and Keith Wyss and Kevin Shih and Kezhi Kong and Khanh Nguyen and Khushi Bhardwaj and Kirill Buryak and Kirthi Shankar Sivamani and Konstantinos Krommydas and Kris Murphy and Krishna C. Puvvada and Krzysztof Pawelec and Kumar Anik and Laikh Tewari and Laya Sleiman and Leo Du and Leon Derczynski and Li Ding and Lilach Ilan and Lingjie Wu and Lizzie Wei and Luis Vega and Lun Su and Maarten Van Segbroeck and Maer Rodrigues de Melo and Magaret Zhang and Mahan Fathi and Makesh Narsimhan Sreedhar and Makesh Sreedhar and Makesh Tarun Chandran and Manuel Reyes Gomez and Maor Ashkenazi and Marc Cuevas and Marc Romeijn and Margaret Zhang and Mark Cai and Mark Gabel and Markus Kliegl and Martyna Patelka and Maryam Moosaei and Matthew Varacalli and Matvei Novikov and Mauricio Ferrato and Mehrzad Samadi and Melissa Corpuz and Meng Xin and Mengdi Wang and Mengru Wang and Meredith Price and Micah Schaffer and Michael Andersch and Michael Boone and Michael Evans and Michael Z Wang and Miguel Martinez and Mikail Khona and Mike Chrzanowski and Mike Hollinger and Mingyuan Ma and Minseok Lee and Mohammad Dabbah and Mohammad Shoeybi and Mostofa Patwary and Nabin Mulepati and Nader Khalil and Najeeb Nabwani and Nancy Agarwal and Nanthini Balasubramaniam and Narimane Hennouni and Narsi Kodukula and Natalie Hereth and Nathaniel Pinckney and Nave Assaf and Negar Habibi and Nestor Qin and Neta Zmora and Netanel Haber and Nick Reamaroon and Nickson Quak and Nidhi Bhatia and Nikhil Jukar and Nikki Pope and Nikolai Ludwig and Nima Tajbakhsh and Nir Ailon and Nirmal Juluru and Nirmalya De and Nowel Pitt and Oleg Rybakov and Oleksii Hrinchuk and Oleksii Kuchaiev and Olivier Delalleau and Oluwatobi Olabiyi and Omer Ullman Argov and Omri Almog and Omri Puny and Oren Tropp and Otavio Padovani and Ouye Xie and Parth Chadha and Pasha Shamis and Paul Gibbons and Pavlo Molchanov and Peter Belcak and Peter Jin and Pinky Xu and Piotr Januszewski and Pooya Jannaty and Prachi Shevate and Pradeep Thalasta and Pranav Prashant Thombre and Prasoon Varshney and Prerana Gambhir and Pritam Gundecha and Przemek Tredak and Qing Miao and Qiyu Wan and Quan Tran Minh and Rabeeh Karimi Mahabadi and Rachel Oberman and Rachit Garg and Rahul Kandu and Raina Zhong and Ran El-Yaniv and Ran Zilberstein and Rasoul Shafipour and Renee Yao and Renjie Pi and Richard Mazzarese and Richard Wang and Rick Izzo and Ridhima Singla and Rima Shahbazyan and Rishabh Garg and Ritika Borkar and Ritu Gala and Riyad Islam and Robert Clark and Robert Hesse and Roger Waleffe and Rohit Varma Kalidindi and Rohit Watve and Roi Koren and Ron Fan and Ruchika Kharwar and Ruisi Cai and Ruoxi Zhang and Russell J. Hewett and Ryan Prenger and Ryan Timbrook and Ryota Egashira and Sadegh Mahdavi and Sagar Singh Ashutosh Joshi and Sahil Modi and Samuel Kriman and Sandeep Pombra and Sanjay Kariyappa and Sanjeev Satheesh and Santiago Pombo and Saori Kaji and Satish Pasumarthi and Saurav Mishra and Saurav Muralidharan and Scott Hara and Sean Narenthiran and Sebastian Rogawski and Seonjin Na and Seonmyeong Bak and Sepehr Sameni and Seth Poulos and Shahar Mor and Shantanu Acharya and Shaona Ghosh Adam Lord and Sharath Turuvekere Sreenivas and Shaun Kotek and Shaya Gharghabi and Shelby Thomas and Sheng-Chieh Lin and Shibani Likhite and Shiqing Fan and Shiyang Chen and Shreya Gopal and Shrimai Prabhumoye and Shubham Pachori and Shubham Toshniwal and Shuo Zhang and Shuoyang Ding and Shyam Renjith and Shyamala Prayaga and Siddhartha Jain and Simeng Sun and Sirisha Rella and Sirshak Das and Smita Ithape and Sneha Harishchandra S and Somshubra Majumdar and Soumye Singhal and Sri Harsha Singudasu and Sriharsha Niverty and Stas Sergienko and Stefana Gloginic and Stefania Alborghetti and Stephen Ge and Stephen McCullough and Sugam Dipak Devare and Suguna Varshini Velury and Sukrit Rao and Sumeet Kumar Barua and Sunny Gai and Suseella Panguluri and Sushil Koundinyan and Swathi Patnam and Sweta Priyadarshi and Swetha Bhendigeri and Syeda Nahida Akter and Sylendran Arunagiri and Tailling Yuan and Talor Abramovich and Tan Bui and Tan Yu and Terry Kong and Thanh Do and Thomas Gburek and Thorgane Marques and Tiffany Moore and Tijmen Blankevoort and Tim Moon and Timothy Ma and Tiyasa Mitra and Tomasz Grzegorzek and Tomer Asida and Tomer Bar Natan and Tomer Keren and Tomer Ronen and Traian Rebedea and Trenton Starkey and Tugrul Konuk and Twinkle Vashishth and Tyler Condensa and Udi Karpas and Ushnish De and Vahid Noorozi and Vahid Noroozi and Vanshil Atul Shah and Veena Vaidyanathan and Venkat Srinivasan and Venmugil Elango and Victor Cui and Vijay Korthikanti and Vikas Mehta and Virginia Adams and Virginia Wu and Vitaly Kurin and Vitaly Lavrukhin and Vladimir Anisimov and Wan Seo and Wanli Jiang and Wasi Uddin Ahmad and Wei Du and Wei Ping and Wei-Ming Chen and Wendy Quan and Wenliang Dai and Wenwen Gao and Will Jennings and William Zhang and Xiaowei Ren and Xiaowen Xin and Xin Li and Yang Yu and Yangyi Chen and Yaniv Galron and Yashaswi Karnati and Yejin Choi and Yev Meyer and Yi-Fu Wu and Yian Zhang and Ying Lin and Yonatan Geifman and Yonggan Fu and Yoshi Suhara and Youngeun Kwon and Yuan Zhang and Yuki Huang and Zach Moshe and Zhilin Wang and Zhiyu Cheng and Zhongbo Zhu and Zhuolin Yang and Zihan Liu and Zijia Chen and Zijie Yan and Zuhair Ahmed},
  year          = {2026},
  eprint        = {2604.12374},
  archiveprefix = {arXiv},
  primaryclass  = {cs.LG},
  url           = {https://arxiv.org/abs/2604.12374}
}

@article{dekoninck2026matharena,
  title         = {Beyond Benchmarks: MathArena as an Evaluation Platform for Mathematics with LLMs},
  author        = {Jasper Dekoninck and Nikola Jovanovi\'{c} and Tim Gehrunger and K\'{a}ri R\"{o}gnvaldsson and Ivo Petrov and Chenhao Sun and Martin Vechev},
  year          = {2026},
  eprint        = {2605.00674},
  archiveprefix = {arXiv},
  primaryclass  = {cs.CL},
  url           = {https://arxiv.org/abs/2605.00674}
}

@software{eval_framework,
  title         = {Aleph Alpha Eval Framework},
  author        = {{Aleph Alpha Research}},
  year          = {2026},
  version       = {x.y.z},
  url           = {https://github.com/Aleph-Alpha-Research/eval-framework}
}

@inproceedings{keskar2017large,
  title         = {On Large-Batch Training for Deep Learning: Generalization Gap and Sharp Minima},
  author        = {Keskar, Nitish Shirish and Mudigere, Dheevatsa and Nocedal, Jorge and Smelyanskiy, Mikhail and Tang, Ping Tak Peter},
  booktitle     = {International Conference on Learning Representations},
  year          = {2017}
}

@inproceedings{dinh2017sharp,
  title         = {Sharp Minima Can Generalize For Deep Nets},
  author        = {Dinh, Laurent and Pascanu, Razvan and Bengio, Samy and Bengio, Yoshua},
  booktitle     = {Proceedings of the 34th International Conference on Machine Learning},
  pages         = {1019--1028},
  volume        = {70},
  series        = {Proceedings of Machine Learning Research},
  publisher     = {PMLR},
  year          = {2017}
}

@inproceedings{izmailov2018averaging,
  title         = {Averaging Weights Leads to Wider Optima and Better Generalization},
  author        = {Izmailov, Pavel and Podoprikhin, Dmitrii and Garipov, Timur and Vetrov, Dmitry P. and Wilson, Andrew Gordon},
  booktitle     = {Proceedings of the Thirty-Fourth Conference on Uncertainty in Artificial Intelligence},
  pages         = {876--885},
  year          = {2018}
}

@inproceedings{foret2021sam,
  title         = {Sharpness-Aware Minimization for Efficiently Improving Generalization},
  author        = {Foret, Pierre and Kleiner, Ariel and Mobahi, Hossein and Neyshabur, Behnam},
  booktitle     = {International Conference on Learning Representations},
  year          = {2021}
}

@inproceedings{liu2023same,
  title         = {Same Pre-training Loss, Better Downstream: Implicit Bias Matters for Language Models},
  author        = {Liu, Hong and Xie, Sang Michael and Li, Zhiyuan and Ma, Tengyu},
  booktitle     = {Proceedings of the 40th International Conference on Machine Learning},
  pages         = {22188--22214},
  volume        = {202},
  series        = {Proceedings of Machine Learning Research},
  publisher     = {PMLR},
  year          = {2023}
}

@inproceedings{li2024forgetting,
  title         = {Revisiting Catastrophic Forgetting in Large Language Model Tuning},
  author        = {Li, Hongyu and Ding, Liang and Fang, Meng and Tao, Dacheng},
  booktitle     = {Findings of the Association for Computational Linguistics: EMNLP 2024},
  pages         = {4297--4308},
  publisher     = {Association for Computational Linguistics},
  year          = {2024},
  doi           = {10.18653/v1/2024.findings-emnlp.249}
}

@inproceedings{yano2026pretraining,
  title         = {Pre-training LLM without Learning Rate Decay Enhances Supervised Fine-Tuning},
  author        = {Yano, Kazuki and Kiyono, Shun and Kobayashi, Sosuke and Takase, Sho and Suzuki, Jun},
  booktitle     = {International Conference on Learning Representations},
  year          = {2026},
  eprint        = {2603.16127},
  archiveprefix = {arXiv}
}

@article{li2025modelmerging,
  title         = {Model Merging in Pre-training of Large Language Models},
  primaryclass  = {cs.CL},
  author        = {Li, Yunshui and Ma, Yiyuan and Yan, Shen and Zhang, Chaoyi and Liu, Jing and Lu, Jianqiao and Xu, Ziwen and Chen, Mengzhao and Wang, Minrui and Zhan, Shiyi and Ma, Jin and Lai, Xunhao and Luo, Yao and Bin, Xingyan and Ren, Hongbin and Han, Mingji and Hao, Wenhao and Yi, Bairen and Liu, LingJun and Ma, Bole and Jia, Xiaoying and Xun, Zhou and Xiang, Liang and Wu, Yonghui},
  journal       = {arXiv preprint arXiv:2505.12082},
  year          = {2025},
  eprint        = {2505.12082},
  archiveprefix = {arXiv}
}

@inproceedings{tian2025wsm,
  title         = {{WSM}: Decay-Free Learning Rate Schedule via Checkpoint Merging for {LLM} Pre-training},
  primaryclass  = {cs.LG},
  author        = {Tian, Changxin and Wang, Jiapeng and Zhao, Qian and Chen, Kunlong and Liu, Jia and Liu, Ziqi and Mao, Jiaxin and Zhao, Wayne Xin and Zhang, Zhiqiang and Zhou, Jun},
  booktitle     = {International Conference on Learning Representations},
  year          = {2026},
  eprint        = {2507.17634},
  archiveprefix = {arXiv}
}

@inproceedings{watts2026sharpness,
  title         = {Sharpness-Aware Pretraining Mitigates Catastrophic Forgetting},
  author        = {Watts, Ishaan and Li, Catherine and Goyal, Sachin and Springer, Jacob Mitchell and Raghunathan, Aditi},
  booktitle     = {Proceedings of the 43rd International Conference on Machine Learning},
  year          = {2026},
  eprint        = {2605.02105},
  archiveprefix = {arXiv},
  primaryclass  = {cs.LG}
}

@article{singh2026trinity,
  title         = {Arcee Trinity Large Technical Report},
  author        = {Singh, Varun and Krauss, Lucas and Jaghouar, Sami and Sirovatka, Matej and Goddard, Charles and Obied, Fares and Ong, Jack Min and Straube, Jannik and Fern and Harley, Aria and Stewart, Conner and Kealty, Colin and Panahi, Maziyar and Kirsten, Simon and Deshpande, Anushka and Vij, Anneketh and Bresnu, Arthur and Veldurthi, Pranav and Ravishankar, Raghav and Bishnoi, Hardik and {DatologyAI Team} and {Arcee AI Team} and {Prime Intellect Team} and McQuade, Mark and Hagemann, Johannes and Atkins, Lucas},
  year          = {2026},
  eprint        = {2602.17004},
  archiveprefix = {arXiv},
  primaryclass  = {cs.LG}
}
\endgroup

\newpage

\appendix

\section{Related Work}
\label{app:related_work}

\paragraph{Checkpoint quality and downstream adaptability.}
A central assumption in model development is that better intermediate metrics imply a better model for subsequent training.
Prior work shows that this need not hold even when pretraining loss is matched.
\citet{liu2023same} find language models with similar pretraining loss but substantially different downstream transfer performance and associate this difference with implicit biases of the pretraining procedure.
More directly, \citet{yano2026pretraining} compare decay-based and decay-free LLM pretraining and find that learning-rate decay can improve pretraining metrics while reducing performance after SFT.
Their result persists with additional mid-training and is associated with sharper pretrained solutions.
Related work on LLM fine-tuning connects loss-landscape geometry to catastrophic forgetting, finding that flatter solutions preserve pretrained capabilities better during adaptation \citep{li2024forgetting}.
Our setting complements these results by following checkpoint rankings through several consecutive training stages and asking when intermediate evaluations become predictive of the final post-SFT ranking.

\paragraph{Flat minima and downstream adaptability.}
The relation between local geometry and model quality has a long history in neural networks.
Flat minima have been associated with improved generalization \citep{keskar2017large}, and methods such as stochastic weight averaging \citep{izmailov2018averaging} and sharpness-aware minimization \citep{foret2021sam} explicitly favor broad low-loss regions.
At the same time, flatness is not an intrinsic property without specifying a parameterization and metric: functionally equivalent networks can have arbitrarily different sharpness under common definitions \citep{dinh2017sharp}.

For language models, \citet{liu2023same} show that pretraining-loss flatness, measured through Hessian-based quantities, correlates with downstream transfer even when pretraining loss itself does not.
\citet{li2024forgetting} connect flatter LLM loss landscapes to reduced forgetting during fine-tuning, while \citet{yano2026pretraining} find that learning-rate decay leads to sharper pretrained solutions together with weaker post-SFT performance.  \citet{watts2026sharpness} show that SAM reduces forgetting after post-training and quantization.
These findings motivate asking whether local geometry contains information about how a checkpoint will behave under subsequent training.

Our solution-density probe is closely related to this literature but measures a different object.
A conventional local flatness measure evaluates the change in the pretraining loss around checkpoint $\theta$.
For example, for $\epsilon\sim\mathcal{N}(0,\sigma^2 I)$, one may consider
\begin{equation}
  F_{\mathrm{pre}}(\theta,\sigma) =
  \mathbb{E}_{\epsilon}
  \left[
    L_{\mathrm{pre}}(\theta+\epsilon)-L_{\mathrm{pre}}(\theta)
    \right].
  \label{eq:pretraining_flatness}
\end{equation}
Under a local second-order approximation,
\begin{equation}
  F_{\mathrm{pre}}(\theta,\sigma)
  \approx
  \frac{\sigma^2}{2}
  \operatorname{Tr}
  \left(
  \nabla^2 L_{\mathrm{pre}}(\theta)
  \right),
  \label{eq:flatness_hessian}
\end{equation}
connecting isotropic perturbation flatness to Hessian-based sharpness.

Our probe instead evaluates whether nearby parameters preserve an external capability.
For benchmark $b$ with score $s_b$, we measure
\begin{equation}
  \delta_{\theta,b}(\tau;\sigma)
  =
  \Pr_{\epsilon\sim\mathcal{N}(0,\sigma^2 I)}
  \left[
    \frac{s_b(\theta+\epsilon)}
    {s_b(\theta)}
    \geq \tau
    \right].
  \label{eq:solution_density_related}
\end{equation}
The two quantities therefore probe the same local parameter neighborhood but apply different functions to it:
\[
  \underbrace{L_{\mathrm{pre}}(\theta+\epsilon)}
  _{\text{pretraining-loss geometry}}
  \qquad\text{vs.
  }\qquad
  \underbrace{s_b(\theta+\epsilon)}
  _{\text{capability retention}}.
\]

There is no general implication between them.
A direction can increase pretraining loss while leaving a benchmark capability unchanged, or preserve pretraining loss while disrupting parameters important for a particular downstream capability.
Moreover, benchmark scores are generally discrete and need not share the local differential geometry of the pretraining objective.
Flatness and solution density may therefore reflect a common robustness of the learned solution, but they are not equivalent measures.

We observe that \constant and \merge have substantially more performance-retaining neighborhoods than \cooldown, consistent with the hypothesis that local geometry relates to subsequent adaptability.
However, our experiment does not show that solution density causes, or generally predicts, better post-training.
We therefore use it as a diagnostic of checkpoint geometry rather than a causal explanation.

\paragraph{Checkpoint averaging and learning-rate decay.}
Checkpoint averaging provides a second connection to the geometry of our three pretraining sources.
Averaging points along an optimization trajectory is well established: stochastic weight averaging, for example, combines checkpoints obtained under sustained or cyclical learning rates and tends to locate broader solutions than the final SGD iterate \citep{izmailov2018averaging}.
Checkpoint merging has since been applied directly during LLM pretraining.
\citet{li2025modelmerging} study merging across dense and MoE models and show that combining checkpoints from constant-learning-rate trajectories can recover much of the benefit of explicit annealing. \citet{tian2025wsm} develop this connection formally and show how checkpoint weights induce an effective decay over the updates along a constant-LR trajectory.
Checkpoint merging has also been adopted in large-scale LLM training systems such as Nemotron Super \citep{nvidia2026nemotron3superopen}.

\section{Model Architecture and Training Details}
\label{app:training_details}

\subsection{Architecture}

The model is a decoder-only Transformer \citep{vaswani2017attention} with 50 layers, of which the first two are dense and the remaining 48 are mixture-of-experts (MoE) layers.
It has approximately 30B total parameters and activates approximately 3B parameters per token.
\Cref{tab:model_architecture} gives the complete architecture.

\begin{table}[htbp]
  \centering
  \small
  \setlength{\tabcolsep}{7pt}
  \caption{Model architecture.}
  \label{tab:model_architecture}
  \begin{tabular}{p{0.48\linewidth}p{0.40\linewidth}}
    \toprule
    \textbf{Component}                    & \textbf{Value}                      \\
    \midrule
    Vocabulary size                       & 96,000                              \\
    Layers                                & 50 total: 2 dense, 48 MoE           \\
    Hidden size                           & 2,048                               \\
    Attention                             & Causal grouped-query attention      \\
    Query / key-value heads               & 32 / 4                              \\
    Head size                             & 128                                 \\
    Query-key normalization               & Per-head RMSNorm                    \\
    Dense MLP hidden size                 & 6,144                               \\
    Expert hidden size                    & 768                                 \\
    MLP activation                        & SwiGLU                              \\
    Routed / shared experts per MoE layer & 128 / 1                             \\
    Experts selected per token            & 8                                   \\
    Routing                               & Token-choice, top-8 sigmoid routing \\
    Pretraining sequence length           & 4,096                               \\
    Position encoding                     & RoPE, base $\theta=500{,}000$       \\
    \bottomrule
  \end{tabular}
\end{table}

Each attention block uses causal grouped-query attention with 32 query heads and four key-value heads \citep{ainslie2023gqa}.
Query and key representations are normalized independently within each head using RMSNorm \citep{zhang2019rmsnorm}.
We use rotary position embeddings (RoPE) with base $\theta=500{,}000$ \citep{su2021roformer}.
Both dense and expert MLPs use SwiGLU activations \citep{shazeer2020glu}.
Each MoE layer contains 128 routed experts and one shared expert.
Token-choice sigmoid routing selects the top eight routed experts for each token.
The shared expert is applied independently of this selection.

\subsection{Pretraining data and objective}

All parameters are randomly initialized with output matrices initialised to zero.
We pretrain with the causal next-token-prediction objective on a broad and diverse document corpus containing English and German data, among other sources.
Documents are packed into 4,096-token sequences.

Pretraining covers approximately 7.508T tokens over 35,800 optimization steps.
The global batch contains 51,200 sequences, or 209.715M tokens per step, using five gradient-accumulation steps across 512 GPUs.
The learning rate is warmed up linearly for 358 steps and then held constant for the remaining 35,442 steps.

\subsection{Optimization}

We partition parameters by role and dimensionality.
The token embedding matrix, all one-dimensional backbone parameters, router weights, and language-model head are optimized with AdamW \citep{loshchilov2019decoupled}.
Their respective learning rates are $0.02916$, $0.02916$, $0.0001139$, and $0.0004556$.
All other two-dimensional backbone parameters are optimized with Nesterov Muon using spectral norm convention at learning rate $0.01458$ \citep{jordan2024muon}.
Thus embeddings and the output head use AdamW despite being matrices, while Muon is reserved for internal two-dimensional backbone parameters.

We apply weight decay $0.0001221$ independently to all decay-eligible parameters.
Embeddings, normalization parameters, and expert-balancing biases are excluded.
The expert-balancing bias is updated separately using SMEBU (Soft-clamped Momentum Expert Bias Updates; \citealp{singh2026trinity}).
During pretraining, an auxiliary load-balancing loss further encourages uniform expert utilization after routing.

\subsection{Continued training and systems}

For capability-focused continued training, the sequence length increases to 8,192 tokens and the global batch decreases to 25,600 sequences, preserving the 209.715M-token batch.
The 100B-token phase therefore spans approximately 477 optimization steps.
This phase uses a 50-step linear warmup followed by a constant learning rate; long-context adaptation and SFT use the same 50-step warmup.
The auxiliary MoE load-balancing loss is disabled during this phase and remains disabled during long-context adaptation.
MoE load remains balanced throughout this stage despite this.

Long-context adaptation and SFT both use 65,536-token sequences.
Long-context adaptation keeps the 209.715M-token global batch, SFT uses a global batch of 384 sequences (25.2M tokens per step), i.e.\ approximately 397 optimization steps for its 10B tokens.
During long-context adaptation the learning rate decays to $1/3$ of its peak; during SFT it decays from its peak to an absolute value of $10^{-5}$, for every SFT learning-rate factor.

Training runs on 512 GPUs using a PyTorch-based TorchTitan distributed-training stack \citep{liang2024torchtitan}.
We train in bfloat16 with optimizer states, gradients, accumulation and master weights in float32.
Gradients are clipped to a maximum global norm of $1.0$.
Optimizer state is reset and learning-rate warmup is repeated at each stage, as described in the main text.

\subsection{\merge construction and effective update weights}
\label{app:merge_details}

\merge does not add optimization steps.
It combines a trailing window from the \constant run using WMA coefficients that rise linearly from the oldest to the newest checkpoint.
For checkpoints $\theta_1,\ldots,\theta_{20}$ ordered from oldest to newest, the selected source is
\begin{equation}
  \theta_{\mathrm{merge}}
  =
  \sum_{i=1}^{20}\frac{i}{210}\theta_i.
\end{equation}
The oldest checkpoint therefore receives weight $1/210$, while the newest receives weight $20/210$.

To expose the effect on individual updates, write $\theta_i=\theta_0+\sum_{t=1}^{i}\Delta\theta_t$, where $\Delta\theta_t=\theta_t-\theta_{t-1}$.
Substitution and exchange of the summations give
\begin{equation}
  \theta_{\mathrm{merge}}
  =
  \theta_0+\sum_{t=1}^{20}q_t\Delta\theta_t,
  \qquad
  q_t=\sum_{i=t}^{20}\frac{i}{210}.
\end{equation}
Although the checkpoint weights increase linearly, their cumulative coefficients on parameter updates decrease across the window: $q_1=1$ for the earliest update and $q_{20}=20/210\approx0.095$ for the latest.
The merge therefore attenuates later updates after training.
This resembles learning-rate decay at the level of the final weighted update history, but it is not equivalent to cosine cooldown because the optimizer never follows the decayed trajectory.

\citet{li2025modelmerging} report that merging constant-learning-rate checkpoints can attain performance comparable to annealed pretraining checkpoints, and \citet{tian2025wsm} formalize model averaging schemes that emulate several decay schedules.
Their experiments use other training runs and model families, so they motivate this source construction rather than validate it for our pipeline.

\section{Evaluation Suites and Aggregate Construction}
\label{app:evaluation_details}

We use a stage-specific evaluation suite at each training boundary.
All evaluations are run with \citep{eval_framework}.
Pre, Mid, and Long use completion-style inference, while SFT uses chat-formatted prompts and the original inference and evaluation adapters.
The suite also expands across the pre-SFT stages: Mid adds long-context evaluation to the Pre clusters, and Long uses a larger code and long-context suite.
\Cref{tab:evaluation_suites} lists every score included in each reported aggregate.

\begin{table}[htbp]
  \centering
  \scriptsize
  \setlength{\tabcolsep}{3.0pt}
  \caption{Stage-specific evaluation suites and aggregation.
    Lengths in parentheses are context lengths.
    Pre, Mid, and Long first average scores within each cluster and then weight the cluster means equally.
    SFT directly averages its six benchmark values.
  }
  \label{tab:evaluation_suites}
  \begin{tabular}{p{0.065\linewidth}p{0.13\linewidth}p{0.565\linewidth}p{0.18\linewidth}}
    \toprule
    \textbf{Stage}                                                     & \textbf{Cluster}   & \textbf{Benchmark scores}                                                                                                                                                                                                     & \textbf{Aggregation}                  \\
    \midrule
    Pre                                                                & General EN         & ARC \citep{clark2018arc}, HellaSwag \citep{zellers2019hellaswag}, MMLU \citep{hendrycks2020mmlu}, MMLU-Pro \citep{wang2024mmlupro}, PIQA \citep{bisk2019piqa}, TriviaQA \citep{joshi2017triviaqa}                             & Equal mean of 3 cluster means         \\
                                                                       & Math EN            & GSM8K \citep{cobbe2021gsm8k}, MATH Minerva \citep{hendrycks2021math}                                                                                                                                                          &                                       \\
                                                                       & Code EN            & HumanEval \citep{chen2021humaneval}                                                                                                                                                                                           &                                       \\
    \midrule
    Mid                                                                & General EN         & ARC, HellaSwag, MMLU, MMLU-Pro, PIQA, TriviaQA                                                                                                                                                                                & Equal mean of 4 cluster means         \\
                                                                       & Math EN            & GSM8K, MATH Minerva                                                                                                                                                                                                           &                                       \\
                                                                       & Code EN            & HumanEval                                                                                                                                                                                                                     &                                       \\
                                                                       & Long Context       & HELMET JSON-KV \citep{yen2024helmet} (8k); RULER \citep{hsieh2024ruler}
    NIAH and QA (4k, 8k, 16k, 32k, 64k); RULER VT and WE (4k, 8k, 16k) &                                                                                                                                                                                                                                                                                            \\
    \midrule
    Long                                                               & General EN         & ARC, HellaSwag, MMLU, MMLU-Pro, PIQA, TriviaQA                                                                                                                                                                                & Equal mean of 4 cluster means         \\
                                                                       & Math EN            & GSM8K, MATH Minerva                                                                                                                                                                                                           &                                       \\
                                                                       & Code EN            & HumanEval, MBPP \citep{austin2021program}                                                                                                                                                                                     &                                       \\
                                                                       & Long Context       & HELMET JSON-KV (8k, 16k, 32k, 64k); RULER NIAH, QA, and VT (4k, 8k, 16k, 32k, 64k); RULER WE (4k, 8k, 16k)                                                                                                                    &                                       \\
    \midrule
    SFT                                                                & Direct task scores & AIME 2026 \citep{dekoninck2026matharena}, DAPO Math \citep{yu2025dapo}, Skywork OR1 Math \citep{he2025skywork}, IFBench \citep{pyatkin2025ifbench}, HumanEval+ \citep{liu2023evalplus}, GPQA Diamond CoT \citep{rein2023gpqa} & Unweighted mean of 6 benchmark values \\
    \bottomrule
  \end{tabular}
\end{table}

For checkpoint $i$, let $s_{i,b}\in[0,1]$ be the scalar score for benchmark configuration $b$, and let $\mu_i(\mathcal{B})$ denote the mean over a set of configurations $\mathcal{B}$.
Let $\mathcal{G}$ and $\mathcal{M}$ denote the General EN and Math EN sets in \cref{tab:evaluation_suites}.
$\mathcal{C}_t$ and $\mathcal{L}_t$ denote the stage-specific Code EN and Long Context sets, and $\mathcal{S}$ denotes the six SFT benchmark values.
The reported stage aggregate is
\begin{equation}
  \begin{aligned}
    \mu_i(\mathcal{B})
     & =\frac{1}{|\mathcal{B}|}\sum_{b\in\mathcal{B}}s_{i,b},
    \\[-1mm]
    A_i^{(t)}
     & =
    \begin{cases}
      \dfrac{\mu_i(\mathcal{G})+\mu_i(\mathcal{M})+\mu_i(\mathcal{C}_{\mathrm{Pre}})}{3},
       & t=\mathrm{Pre},                     \\[2mm]
      \dfrac{\mu_i(\mathcal{G})+\mu_i(\mathcal{M})+\mu_i(\mathcal{C}_{t})+\mu_i(\mathcal{L}_{t})}{4},
       & t\in\{\mathrm{Mid},\mathrm{Long}\}, \\[2mm]
      \dfrac{1}{6}\displaystyle\sum_{b\in\mathcal{S}}s_{i,b},
       & t=\mathrm{SFT}.
    \end{cases}
  \end{aligned}
  \label{eq:stage_aggregate}
\end{equation}
Consequently, individual pre-SFT benchmarks do not all have equal final weight.
For example, Pre contains nine benchmark scores, but HumanEval alone forms the Code EN cluster and therefore receives one third of the aggregate rather than one ninth.
Mid and Long similarly assign one quarter of the aggregate to each cluster, irrespective of the number of scores in that cluster.
Each listed long-context task--length pair is one score when forming its Long Context cluster mean.

The SFT aggregate is instead a direct unweighted mean of six benchmark values.
IFBench contributes one value, formed by averaging its loose and strict prompt-level scores.
This single value enters the six-task mean.
We never pool example-level correct counts across benchmarks.
Invalid or unparsable answers are scored according to the original adapter and are not removed before averaging.

For HumanEval+, the primary score is pass@1 after extracting the final fenced Python block, matching the original adapter.
\Cref{app:code_audit} reports the diagnostic first-block rescore, which is not substituted into any primary aggregate. \Cref{tab:source_profiles} gives all six post-SFT benchmark scores and their aggregate for the \constant, \cooldown, and \merge comparison. \Cref{tab:fixed_sft3_branches_app} gives the corresponding intermediate and post-SFT aggregates for every training trajectory used in the correlation analysis.

\section{Complete Training Trajectories}
\label{app:branch_data}

\Cref{tab:fixed_sft3_branches_app} lists the training trajectories used in the stage-association analysis.
Every post-SFT aggregate in that analysis comes from a checkpoint trained with SFT learning-rate factor $1/3$.
No rate is selected per trajectory.
Separately, each of the nine \cooldown trajectories receives an auxiliary SFT learning-rate sweep, summarized in \cref{app:sft_rates}.

\begin{table}[htbp]
  \centering
  \small
  \caption{Intermediate and post-SFT aggregate scores.
    All eleven trajectories enter the reported correlations.
  }
  \label{tab:fixed_sft3_branches_app}
  \begin{tabular}{llrrr}
    \toprule
    \textbf{Source} & \textbf{Mid, Long factors} & \textbf{Mid aggregate} & \textbf{Long aggregate} & \textbf{Post-SFT aggregate} \\
    \midrule
    \cooldown       & $1,1$                      & 0.515                  & 0.637                   & 0.247                       \\
    \cooldown       & $1,1/3$                    & 0.564                  & 0.626                   & 0.199                       \\
    \cooldown       & $1,1/9$                    & 0.544                  & 0.621                   & 0.187                       \\
    \cooldown       & $1/3,1$                    & 0.557                  & 0.626                   & 0.228                       \\
    \cooldown       & $1/3,1/3$                  & 0.543                  & 0.618                   & 0.191                       \\
    \cooldown       & $1/3,1/9$                  & 0.537                  & 0.610                   & 0.119                       \\
    \cooldown       & $1/9,1$                    & 0.516                  & 0.602                   & 0.160                       \\
    \cooldown       & $1/9,1/3$                  & 0.521                  & 0.588                   & 0.104                       \\
    \cooldown       & $1/9,1/9$                  & 0.512                  & 0.581                   & 0.087                       \\
    \constant       & $1,1$                      & 0.556                  & 0.636                   & 0.360                       \\
    \merge          & $1,1$                      & 0.540                  & 0.644                   & 0.363                       \\
    \bottomrule
  \end{tabular}
\end{table}

\begin{table}[htbp]
  \centering
  \small
  \caption{Association between intermediate and post-SFT aggregate scores across the eleven training trajectories.
    Reported $p$-values are two-sided.
  }
  \label{tab:stage_correlations}
  \begin{tabular}{lrrrr}
    \toprule
    \textbf{Signal}        & \textbf{Pearson $r$} & \textbf{$p$} & \textbf{Spearman $\rho$} & \textbf{$p$} \\
    \midrule
    Mid-training aggregate & 0.473                & 0.142        & 0.482                    & 0.133        \\
    Long-context aggregate & 0.884                & $<0.001$     & 0.964                    & $<0.001$     \\
    \bottomrule
  \end{tabular}
\end{table}

The correlations are computed from the unrounded aggregate scores using two-sided tests.
Pearson coefficients have 95\% confidence intervals of $[-0.177,0.836]$ after mid-training and $[0.605,0.970]$ after long-context adaptation.
Spearman coefficients use average ranks.
Restricting the analysis to the nine \cooldown trajectories yields mid-training $r=0.448$ ($p=0.226$) and $\rho=0.450$ ($p=0.224$), compared with long-context $r=0.935$ ($p<0.001$) and $\rho=0.950$ ($p<0.001$).

\section{Stopping and Repetition Audit}
\label{app:stopping_audit}

We compare retained raw generations from the fixed-recipe \cooldown checkpoint with the corresponding \constant and \merge checkpoints.
After observing severe repetition, we additionally inspect \cooldown $(1,1)$ with SFT factor $1$, which has the highest post-SFT aggregate among the complete \cooldown checkpoints in the recorded SFT-rate sweep.

The audit covers 294 IFBench prompts, 240 AIME generations, and 198 GPQA Diamond prompts per checkpoint under a 32,768-token payload cap.
AIME and GPQA record generated sequence length directly.
IFBench is retokenized with the tokenizer used for training.
``near cap'' denotes at least 32,000 tokens.

\begin{table}[htbp]
  \centering
  \scriptsize
  \setlength{\tabcolsep}{3.2pt}
  \caption{Stopping behavior under the 32,768-token cap.
    Cells report median tokens followed by the percentage near the cap for IFBench or exactly at the cap for AIME and GPQA.
  }
  \label{tab:stopping_lengths}
  \begin{tabular}{lrrr}
    \toprule
    \textbf{Checkpoint}          & \textbf{IFBench} & \textbf{AIME}  & \textbf{GPQA}   \\
    \midrule
    \cooldown $(1,1)$, SFT $1/3$ & 32,616 / 75.9\%  & 32,768 / 100\% & 32,768 / 99.0\% \\
    \constant $(1,1)$, SFT $1/3$ & 1,899 / 33.0\%   & 6,552 / 29.6\% & 32,768 / 54.5\% \\
    \merge $(1,1)$, SFT $1/3$    & 1,805 / 33.3\%   & 5,135 / 25.8\% & 32,768 / 54.0\% \\
    \midrule
    \cooldown $(1,1)$, SFT $1$   & 32,676 / 93.2\%  & 32,768 / 100\% & 32,768 / 99.0\% \\
    \bottomrule
  \end{tabular}
\end{table}

The fixed-recipe \cooldown checkpoint is substantially less likely to terminate than the \constant and \merge checkpoints.
Median zlib-to-raw byte ratios are 0.019, 0.044, and 0.027 on IFBench, AIME, and GPQA, compared with 0.288, 0.310, and 0.059 for \constant and 0.299, 0.328, and 0.066 for \merge.
The long \cooldown outputs therefore contain extensive repeated material.

Changing \cooldown from SFT factor $1/3$ to factor $1$ does not repair the behavior.
The corresponding compression ratios remain 0.022, 0.053, and 0.031.

\subsection{SFT-rate diagnostic}
\label{app:sft_rates}

The additional \cooldown checkpoints are used only as a debugging sweep.
All three SFT factors are complete for all nine mid/long trajectories, giving 27 evaluated checkpoints.
These additional checkpoints are not included in the primary source comparison or trajectory correlations.

\begin{table}[htbp]
  \centering
  \small
  \caption{Auxiliary \cooldown diagnostic by SFT learning-rate factor.
    Values are descriptive averages across complete runs.
    Reasoning is the mean of AIME, DAPO, Skywork, and GPQA.
  }
  \label{tab:sft_factor}
  \begin{tabular}{lrrrr}
    \toprule
    \textbf{SFT factor} & $\boldsymbol{n}$ & \textbf{Post-SFT aggregate} & \textbf{Reasoning} & \textbf{HumanEval+} \\
    \midrule
    $1$                 & 9                & 0.190                       & 0.187              & 0.168               \\
    $1/3$               & 9                & 0.169                       & 0.140              & 0.243               \\
    $1/9$               & 9                & 0.153                       & 0.109              & 0.293               \\
    \bottomrule
  \end{tabular}
\end{table}

On the fixed $(1,1)$ upstream trajectory, changing the SFT factor from $1/3$ to $1$ increases the post-SFT aggregate from 0.247 to 0.294, the highest recorded complete \cooldown result.
Its original HumanEval+ score is 0.067.

Across upstream schedules, lower SFT learning rates improve HumanEval+ on average while reducing the reasoning and overall aggregates.
No tested SFT factor therefore dominates all recorded capabilities.
Final SFT loss is also not a sufficient selection signal: the lowest-loss checkpoint is the extraction-sensitive \cooldown $(1,1)$, SFT-factor-$1$ checkpoint rather than a uniformly superior assistant.

\subsection{Compact-answer sensitivity}

\begin{table}[htbp]
  \centering
  \scriptsize
  \setlength{\tabcolsep}{4.5pt}
  \caption{Official accuracy and a heuristic first-explicit-answer rescore on identical generations.
    The heuristic compares the earliest explicit or boxed answer with the target and is used only diagnostically.
  }
  \label{tab:first_answer}
  \begin{tabular}{lrr}
    \toprule
    \textbf{Checkpoint}          & \textbf{AIME official $\rightarrow$ first} & \textbf{GPQA official $\rightarrow$ first} \\
    \midrule
    \cooldown $(1,1)$, SFT $1/3$ & 17.08\% $\rightarrow$ 20.42\%              & 20.71\% $\rightarrow$ 22.73\%              \\
    \cooldown $(1,1)$, SFT $1$   & 18.33\% $\rightarrow$ 23.75\%              & 25.25\% $\rightarrow$ 29.80\%              \\
    \constant $(1,1)$, SFT $1/3$ & 22.08\% $\rightarrow$ 22.08\%              & 23.74\% $\rightarrow$ 23.74\%              \\
    \merge $(1,1)$, SFT $1/3$    & 21.67\% $\rightarrow$ 21.25\%              & 22.22\% $\rightarrow$ 24.24\%              \\
    \bottomrule
  \end{tabular}
\end{table}

First-answer sensitivity is measurable for \cooldown but much smaller than the HumanEval+ first-block effect.
Official GPQA parsing marks 115/198 and 106/198 \cooldown responses invalid, but also 111/198 \constant and 114/198 \merge responses, so GPQA extraction fragility is not specific to \cooldown.

IFBench strict scores likewise do not directly expose the stopping failure: \cooldown with SFT factors $1/3$ and $1$ scores 23.47\% and 22.79\%, compared with 22.11\% for the corresponding \constant and \merge checkpoints.
The same response pathology therefore interacts differently with different evaluation contracts.

\subsection{Qualitative examples}

The following examples illustrate, rather than estimate, the observed repetition pattern.

An IFBench response initially satisfies exact keyword-count constraints and subsequently repeats ``(Answer ready.)'' 575 times and its keyword audit 685 times, causing the response to violate the same constraints it had already satisfied.

One AIME generation repeats ``I hope it is correct.'' 4,208 times, reaches the generation cap, and drifts from an earlier correct answer of 190 to an officially extracted answer of 290.

One GPQA response repeats the sentence ``the nitro group is attached to the carbon bearing the nitro group'' 2,012 times, reaches the cap mid-word, and is marked invalid despite stating the correct option earlier.

The most extreme 32k HumanEval+ response contains 5,318 fenced code blocks but only four unique blocks.
One block appears 5,314 times.

These examples show why one underlying stopping failure can produce different measured penalties under whole-response, compact-answer, and final-block evaluators.

\section{HumanEval+ Re-execution Audit}
\label{app:code_audit}

For response $y$, extractor $E$, and functional verifier $V$, the measured code score is
\begin{equation}
  S_E
  =
  \mathbb{E}[V(E(y))].
\end{equation}

The original evaluation adapter extracts the final fenced Python block.
We compare it with the first fenced block.
Selecting the first syntactically valid block produces the same aggregate as selecting the first fenced block in every audited run.

\begin{table}[htbp]
  \centering
  \scriptsize
  \setlength{\tabcolsep}{3.5pt}
  \caption{HumanEval+ pass@1 under alternative extraction from identical generations.
    Counts are out of 164.
  }
  \label{tab:extraction_audit}
  \begin{tabular}{llrrr}
    \toprule
    \textbf{Checkpoint}          & \textbf{Cap} & \textbf{Last block} & \textbf{First block} & \textbf{$\Delta$} \\
    \midrule
    \cooldown $(1,1)$, SFT $1$   & 1k           & 9 (5.49\%)          & 97 (59.15\%)         & +53.66            \\
    \cooldown $(1,1)$, SFT $1$   & 32k          & 12 (7.32\%)         & 100 (60.98\%)        & +53.66            \\
    \constant $(1,1)$, SFT $1/3$ & 32k          & 108 (65.85\%)       & 106 (64.63\%)        & $-1.22$           \\
    \merge $(1,1)$, SFT $1/3$    & 32k          & 108 (65.85\%)       & 107 (65.24\%)        & $-0.61$           \\
    \bottomrule
  \end{tabular}
\end{table}

All scores reuse the same raw generations.
Candidate programs are executed against the original 164 HumanEval+ test suites in a network-disabled container.
Re-executing the final block reproduces every original verdict, so extraction is the only changed scoring variable.

For both audited \cooldown caps, no last-block success becomes a first-block failure.
Every gain comes from a response containing a passing early program and a failing final program.
At 32k, \cooldown has a median response length of 32,690 tokens and a 72.6\% near-cap rate, compared with medians of 657 and 484 tokens and near-cap rates of 15.2\% and 7.3\% for \constant and \merge.

\cooldown also has a median of 1,609 fenced blocks per response, with a maximum of 5,318, compared with a median of one and maxima of four and two for \constant and \merge.

Increasing the output cap does not repair the behavior.
\cooldown's first-block accuracy remains near 60\% while its generations continue to contain thousands of repeated blocks.

The fixed-recipe \cooldown SFT-$1/3$ checkpoint has an original HumanEval+ score of 8/164 (4.88\%), but its evaluation artifact retains only aggregate scores.
First-block rescoring would require raw generations that are no longer available.
We therefore do not transfer the corrected score from SFT factor $1$ to the primary source comparison.

\section{Solution Density Probe Details}
\label{app:solution_density}

\subsection{Estimator}

For checkpoint $\theta$, benchmark $b$, threshold $\tau$, and $n_{\theta,b}$ valid perturbations, we estimate
\begin{equation}
  \widehat{\delta}_{\theta,b}(\tau)
  =
  \frac{1}{n_{\theta,b}}
  \sum_{i=1}^{n_{\theta,b}}
  \mathbf{1}
  \left[
    \frac{s_b(\theta+\epsilon_i)}{s_b(\theta)}
    \geq
    \tau
    \right].
\end{equation}

Each score is divided by the unperturbed score from the same checkpoint.
We retain the latest result for each matched perturbation job and exclude results without numeric benchmark summaries.
Missing values are not imputed.
Both perturbation scales contain 100 perturbations per checkpoint and benchmark.

\subsection{Primary perturbation scale}

\begin{figure}[htbp]
  \centering
  \begin{tikzpicture}
  \begin{groupplot}[
      paper axis,
      group style={group size=2 by 1, horizontal sep=0.10\linewidth},
      width=0.45\linewidth,
      height=0.29\linewidth,
      xmin=50,
      xmax=110,
      ymin=0,
      ymax=104,
      xtick={50,70,90,100,110},
      ytick={0,25,50,75,100},
      xlabel={Threshold $\tau$ (\% of unperturbed score)},
    ]
    \nextgroupplot[
      title={(a) GSM8K},
      ylabel={Solution density $\widehat{\delta}(\tau)$ (\%)},
      legend to name=primaryretentionlegend,
      legend columns=3,
      legend style={draw=none, font=\scriptsize, column sep=0.65em},
    ]
    \addplot+[constantcolor, mark=*] coordinates {
        (50,100.0) (55,100.0) (60,100.0) (65,100.0) (70,100.0) (75,93.0)
        (80,80.0) (85,62.0) (90,27.0) (95,10.0) (100,2.0) (105,1.0) (110,0.0)
      };
    \addlegendentry{\constant}
    \addplot+[cooldowncolor, mark=square*] coordinates {
        (50,97.0) (55,93.0) (60,81.0) (65,67.0) (70,51.0) (75,23.0)
        (80,8.0) (85,1.0) (90,0.0) (95,0.0) (100,0.0) (105,0.0) (110,0.0)
      };
    \addlegendentry{\cooldown}
    \addplot+[mergecolor, mark=triangle*] coordinates {
        (50,100.0) (55,100.0) (60,100.0) (65,100.0) (70,97.0) (75,95.0)
        (80,75.0) (85,45.0) (90,13.0) (95,0.0) (100,0.0) (105,0.0) (110,0.0)
      };
    \addlegendentry{\merge}

    \nextgroupplot[title={(b) MBPP}, yticklabels={}]
    \addplot+[constantcolor, mark=*, forget plot] coordinates {
        (50,100.0) (55,100.0) (60,100.0) (65,96.0) (70,92.0) (75,78.0)
        (80,52.0) (85,15.0) (90,3.0) (95,1.0) (100,0.0) (105,0.0) (110,0.0)
      };
    \addplot+[cooldowncolor, mark=square*, forget plot] coordinates {
        (50,95.0) (55,89.0) (60,78.0) (65,53.0) (70,26.0) (75,4.0)
        (80,1.0) (85,0.0) (90,0.0) (95,0.0) (100,0.0) (105,0.0) (110,0.0)
      };
    \addplot+[mergecolor, mark=triangle*, forget plot] coordinates {
        (50,99.0) (55,99.0) (60,99.0) (65,95.0) (70,88.0) (75,74.0)
        (80,42.0) (85,13.0) (90,0.0) (95,0.0) (100,0.0) (105,0.0) (110,0.0)
      };
  \end{groupplot}
\end{tikzpicture}
\pgfplotslegendfromname{primaryretentionlegend}
   \caption{Solution-density profiles at $\sigma_\epsilon=0.005$.
    Each curve reports the fraction of 100 perturbations retaining at least threshold $\tau$ of the corresponding unperturbed score.
  }
  \label{fig:solution_density_primary}
\end{figure}
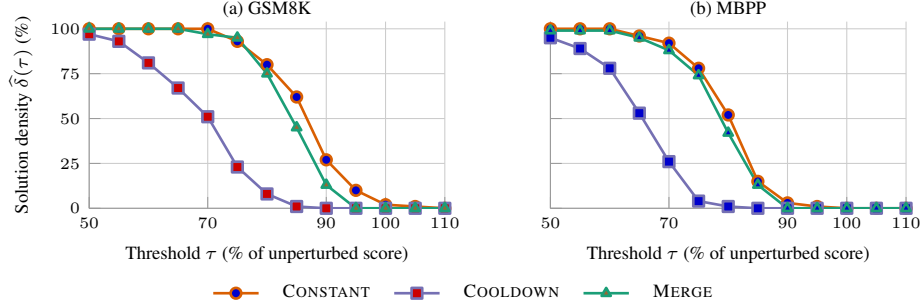

\begin{table}[htbp]
  \centering
  \scriptsize
  \caption{Perturbations above each score threshold at $\sigma_\epsilon=0.005$.
    Cells report counts out of 100 followed by percentages.
  }
  \label{tab:retention_counts}
  \begin{tabular}{llcccccc}
    \toprule
    \textbf{Benchmark} & \textbf{Checkpoint} & $\boldsymbol{n}$ & $\boldsymbol{\tau\geq0.90}$ & $\boldsymbol{\tau\geq0.95}$ & $\boldsymbol{\tau\geq1.00}$ & $\boldsymbol{\tau\geq1.05}$ & $\boldsymbol{\tau\geq1.10}$ \\
    \midrule
    GSM8K              & \constant           & 100              & 27/100 (27.0\%)             & 10/100 (10.0\%)             & 2/100 (2.0\%)               & 1/100 (1.0\%)               & 0/100 (0.0\%)               \\
                       & \cooldown           & 100              & 0/100 (0.0\%)               & 0/100 (0.0\%)               & 0/100 (0.0\%)               & 0/100 (0.0\%)               & 0/100 (0.0\%)               \\
                       & \merge              & 100              & 13/100 (13.0\%)             & 0/100 (0.0\%)               & 0/100 (0.0\%)               & 0/100 (0.0\%)               & 0/100 (0.0\%)               \\
    \midrule
    MBPP               & \constant           & 100              & 3/100 (3.0\%)               & 1/100 (1.0\%)               & 0/100 (0.0\%)               & 0/100 (0.0\%)               & 0/100 (0.0\%)               \\
                       & \cooldown           & 100              & 0/100 (0.0\%)               & 0/100 (0.0\%)               & 0/100 (0.0\%)               & 0/100 (0.0\%)               & 0/100 (0.0\%)               \\
                       & \merge              & 100              & 0/100 (0.0\%)               & 0/100 (0.0\%)               & 0/100 (0.0\%)               & 0/100 (0.0\%)               & 0/100 (0.0\%)               \\
    \bottomrule
  \end{tabular}
\end{table}

Zero cells indicate that no success was observed among the 100 perturbations, not that the underlying probability is zero.
We therefore treat the curves as empirical profiles rather than estimated continuous densities.

\subsection{Smaller perturbation scale}

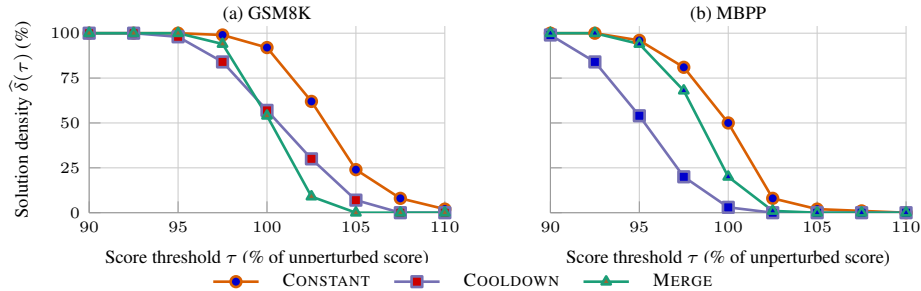
\begin{figure}[htbp]
  \centering
  \begin{tikzpicture}
  \begin{groupplot}[
      paper axis,
      group style={group size=2 by 1, horizontal sep=0.10\linewidth},
      width=0.45\linewidth,
      height=0.29\linewidth,
      xmin=90,
      xmax=110,
      ymin=0,
      ymax=104,
      xtick={90,95,100,105,110},
      ytick={0,25,50,75,100},
      xlabel={Score threshold $\tau$ (\% of unperturbed score)},
    ]
    \nextgroupplot[
      title={(a) GSM8K},
      ylabel={Solution density $\widehat{\delta}(\tau)$ (\%)},
      legend to name=smallretentionlegend,
      legend columns=3,
      legend style={draw=none, font=\scriptsize, column sep=0.65em},
    ]
    \addplot+[constantcolor, mark=*] coordinates {
        (90,100.0) (92.5,100.0) (95,100.0) (97.5,99.0)
        (100,92.0) (102.5,62.0) (105,24.0) (107.5,8.0) (110,2.0)
      };
    \addlegendentry{\constant}
    \addplot+[cooldowncolor, mark=square*] coordinates {
        (90,100.0) (92.5,100.0) (95,98.0) (97.5,84.0)
        (100,57.0) (102.5,30.0) (105,7.0) (107.5,0.0) (110,0.0)
      };
    \addlegendentry{\cooldown}
    \addplot+[mergecolor, mark=triangle*] coordinates {
        (90,100.0) (92.5,100.0) (95,100.0) (97.5,94.0)
        (100,54.0) (102.5,9.0) (105,0.0) (107.5,0.0) (110,0.0)
      };
    \addlegendentry{\merge}

    \nextgroupplot[
      title={(b) MBPP},
      yticklabels={},
    ]
    \addplot+[constantcolor, mark=*, forget plot] coordinates {
        (90,100.0) (92.5,100.0) (95,96.0) (97.5,81.0)
        (100,50.0) (102.5,8.0) (105,2.0) (107.5,1.0) (110,0.0)
      };
    \addplot+[cooldowncolor, mark=square*, forget plot] coordinates {
        (90,99.0) (92.5,84.0) (95,54.0) (97.5,20.0)
        (100,3.0) (102.5,0.0) (105,0.0) (107.5,0.0) (110,0.0)
      };
    \addplot+[mergecolor, mark=triangle*, forget plot] coordinates {
        (90,100.0) (92.5,100.0) (95,94.0) (97.5,68.0)
        (100,20.0) (102.5,1.0) (105,0.0) (107.5,0.0) (110,0.0)
      };
  \end{groupplot}
\end{tikzpicture}

\vspace{-0.6em}
\pgfplotslegendfromname{smallretentionlegend}
   \caption{Solution-density profiles at $\sigma_\epsilon=0.001$.
    \constant has the greatest density around the unperturbed score.
    On MBPP, 3\% of \cooldown perturbations match or exceed the unperturbed score, compared with 50\% for \constant and 20\% for \merge.
  }
  \label{fig:solution_density_small}
\end{figure}

\begin{table}[htbp]
  \centering
  \scriptsize
  \caption{Results at $\sigma_\epsilon=0.001$.
    Mean is the average perturbed score as a percentage of the corresponding unperturbed score.
  }
  \label{tab:retention_counts_small}
  \begin{tabular}{llcccccc}
    \toprule
    \textbf{Benchmark} & \textbf{Checkpoint} & $\boldsymbol{n}$ & \textbf{Mean} & $\boldsymbol{\tau\geq0.95}$ & $\boldsymbol{\tau\geq1.00}$ & $\boldsymbol{\tau\geq1.05}$ & $\boldsymbol{\tau\geq1.10}$ \\
    \midrule
    GSM8K              & \constant           & 100              & 103.5\%       & 100/100 (100.0\%)           & 92/100 (92.0\%)             & 24/100 (24.0\%)             & 2/100 (2.0\%)               \\
                       & \cooldown           & 100              & 100.7\%       & 98/100 (98.0\%)             & 57/100 (57.0\%)             & 7/100 (7.0\%)               & 0/100 (0.0\%)               \\
                       & \merge              & 100              & 100.1\%       & 100/100 (100.0\%)           & 54/100 (54.0\%)             & 0/100 (0.0\%)               & 0/100 (0.0\%)               \\
    \midrule
    MBPP               & \constant           & 100              & 99.7\%        & 96/100 (96.0\%)             & 50/100 (50.0\%)             & 2/100 (2.0\%)               & 0/100 (0.0\%)               \\
                       & \cooldown           & 100              & 95.2\%        & 54/100 (54.0\%)             & 3/100 (3.0\%)               & 0/100 (0.0\%)               & 0/100 (0.0\%)               \\
                       & \merge              & 100              & 98.3\%        & 94/100 (94.0\%)             & 20/100 (20.0\%)             & 0/100 (0.0\%)               & 0/100 (0.0\%)               \\
    \bottomrule
  \end{tabular}
\end{table}

Values above 100\% indicate that a perturbed checkpoint outscored its single unperturbed evaluation.
They should not be interpreted as established local improvements without repeated unperturbed evaluations.
The stable observation is relative: \cooldown has lower solution density around performance-preserving thresholds than \constant and \merge, particularly on MBPP.

\end{document}